\documentclass{artificial-life}

\usepackage[style=apa,natbib=true]{biblatex}
\usepackage{tabularx}
\usepackage{multirow}
\usepackage{amsmath}
\usepackage{amssymb}
\title{Evolving Symbiosis, from Barricelli’s Legacy to Collective Intelligence: a simulated and conceptual approach}

\auth{
    James Ashford \affil{1},
    Marko Cvjetko \affil{2},
    Richard J.G. L{\"o}ffler \affil{3},
    Berfin Sakallioglu \affil{4},
    Alessandro Valerio \affil{5},
    Marta Tataryn \affil{3},
    Benedikt Hartl \affil{6},
    Léo Pio-Lopez \affil{6},
    Stefano Nichele \affil{7}

}

\corresponding{Stefano Nichele}{stefano.nichele@hiof.no}

\affiliations{
    \item Independent Researcher, Oxford, UK
    \item Inria Center at the University of Bordeaux, Bordeaux, France
    \item University of Copenhagen, Copenhagen, Denmark
    \item University of Trieste, Trieste, Italy
    \item Gran Sasso Science Institute, L'Aquila, Italy 
    \item Allen Discovery Center at Tufts University, Medford, USA
    \item Østfold University College, Halden, Norway
}

\abs{
This report documents the work of our group (named SymBa) at the \href{https://aliceworkshop.org/}{ALICE 2026 workshop in Copenhagen}. Inspired by the pioneering work by Nils Aall Barricelli on symbiogenesis of numerical organisms (i.e., 1D cellular automata) in 1953 (70+ years ago!!), we discussed the role of symbiogenesis as mechanism contributing to the origins of life, open-endedness, and collective intelligence. We report replications of Barricelli's original work in 1D worlds, an extension to 2D symbioorganisms, and preliminary experimentation with DNA-norms. We discuss the implications of symbiogenesis for artifical life and artificial intelligence, and outline several opportunities for future works, both at the conceptual level as well as using different substrates (neural networks, neural cellular automata, etc.)
}


\begin{document}

\coverpage           %
\doublespacing       %


\section{Introduction}

\begin{figure}[t]
    \centering
    \includegraphics[width=0.7\textwidth]{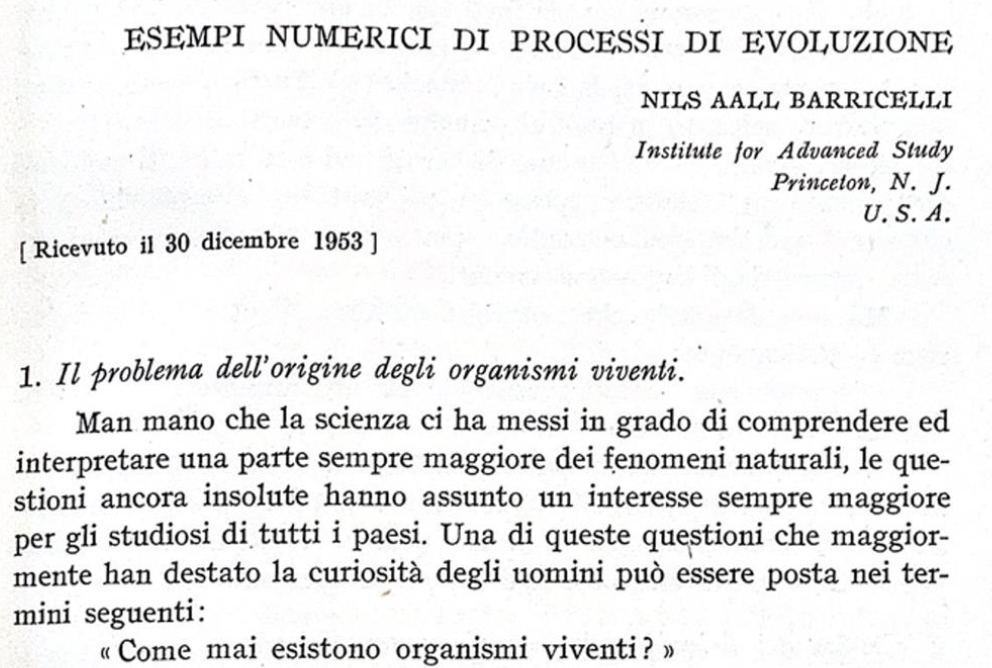}
    \caption{Extract from  original paper by Barricelli published in Italian \citep{barricelli1954esempi}.}
    \label{fig:bar53text}
\end{figure}

Why do living organisms exist? This question was addressed in 1954 by Nils Aall Barricelli \citep{barricelli1954esempi} (see scanned snapshot of the original paper in Fig. \ref{fig:bar53text}), one of the founding fathers of artificial life, using numeric symbioorganisms (one dimensional cellular automata). In his Cellular Automata (CA) models, reproduction and mutation were not sufficient to explain the origin of an evolutionary process (and therefore the origins of life). He proposed that the missing ingredient was symbiogenesis, the creation of a new entity out of a mutually beneficial relationship between two pre-existing entities. Barricelli’s work has not been fully appreciated, however it is still very relevant today. In this paper we will review Barricelli’s ideas on symbiogenesis, recent ideas, and identify future directions and open questions.

\textbf{Opening -- Two original questions to be explored:}
\begin{enumerate}
    \item When do symbiotic associations become more permanent, and what is the benefit/motivation of that?! This is an important ingredient for explaining origin of life, i.e., kickstart the emergence of replicators. 
    \item Is symbiogenesis benificial (or even required) for open-endedness?
\end{enumerate}


\textbf{Central motivation for this report}: 
\begin{itemize}
    \item Biology showed that replication and collaboration works with physics/chemistry.
    \item Barricelli showed in 1953 that it also works with math (numerical organisms). 
    \item  Recently, the importance of symbiotic approaches to (artificial) intelligence and (a)life has been undergoing a renaissance (see, e.g., "Computational Life: How Well-formed, Self-replicating Programs Emerge from Simple Interaction"  \citep{alakuijala2024computational}, or more comprehensively "What is Intelligence?" \citep{Aguera2025WhatIsLife})
    \item We thus ask: How does symbiogenesis relate to our modern scientific and technological views (spanning Machine Learning, generative AI, collective intelligence, diverse intelligence, etc.)? And how does this relate to our improved understanding of biology (see, e.g.,~\cite{Fields2022Competency, Fields2020ScaleFree, Levin2023DAM, levin_technological_2022, McMillen2024, Levin2026})?
\end{itemize}

\section{Historical Ideas of Barricelli}

\begin{figure}[t]
    \centering
    \includegraphics[width=0.6\textwidth]{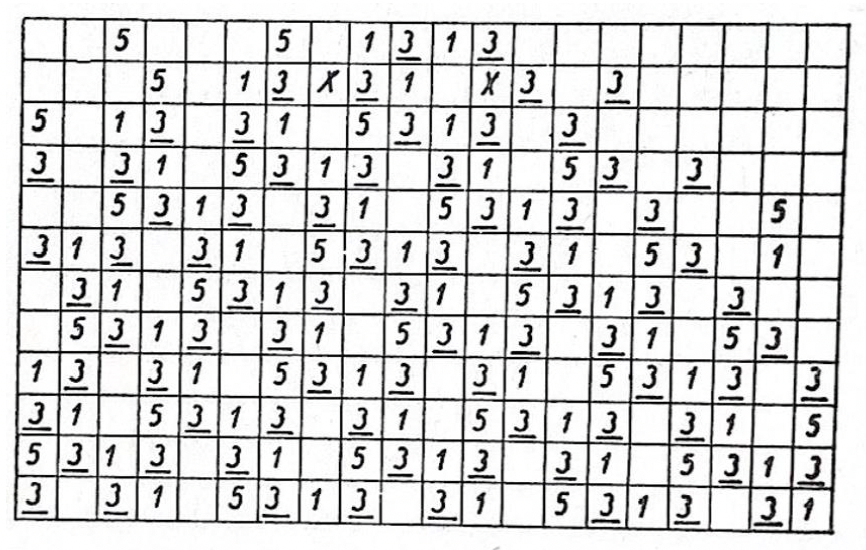}
    \caption{An emergent symbioorganism from Barricelli's work.}
    \label{fig:bar57organism}
\end{figure}

\begin{figure}[t]
    \centering
    \includegraphics[width=0.4\textwidth]{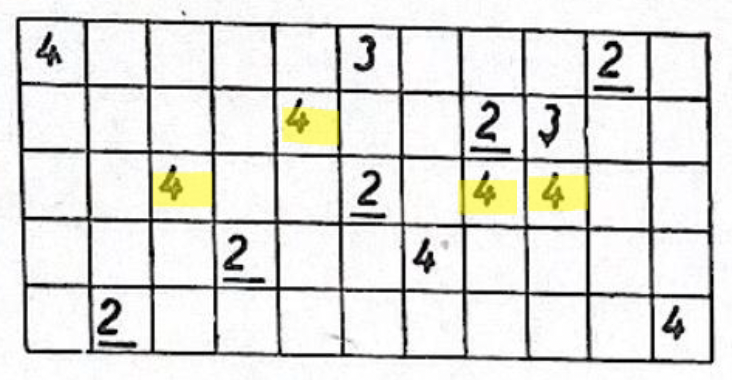}
    \caption{Example of symbiotic interactions causing mutiple replications that would not have happened without symbiosis.}
    \label{fig:barsymbiosis}
\end{figure}

\begin{figure}[t]
    \centering
    \includegraphics[width=0.9\textwidth]{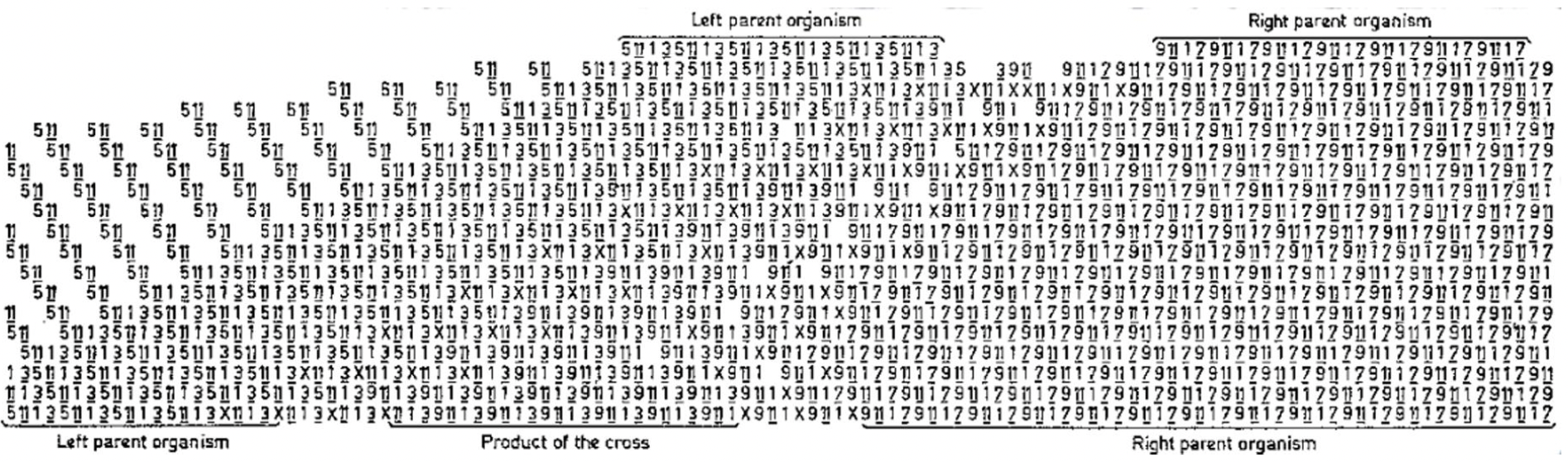}
    \caption{Example of emergent crossover.}
    \label{fig:barcross}
\end{figure}

\begin{figure}[t]
    \centering
    \includegraphics[width=0.4\textwidth]{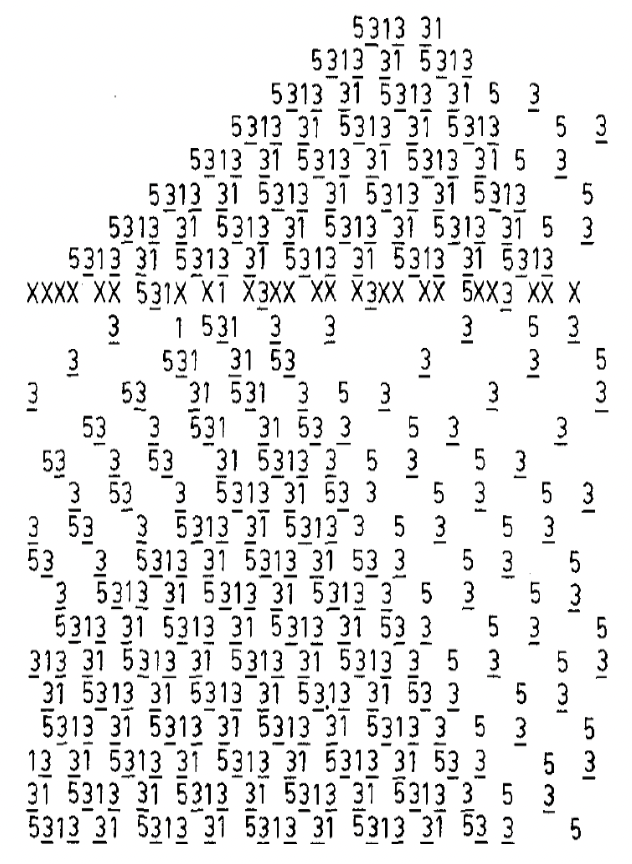}
    \caption{Example of emergent crossover.}
    \label{fig:barrepair}
\end{figure}

\begin{figure}[t]
    \centering
    \includegraphics[width=0.6\textwidth]{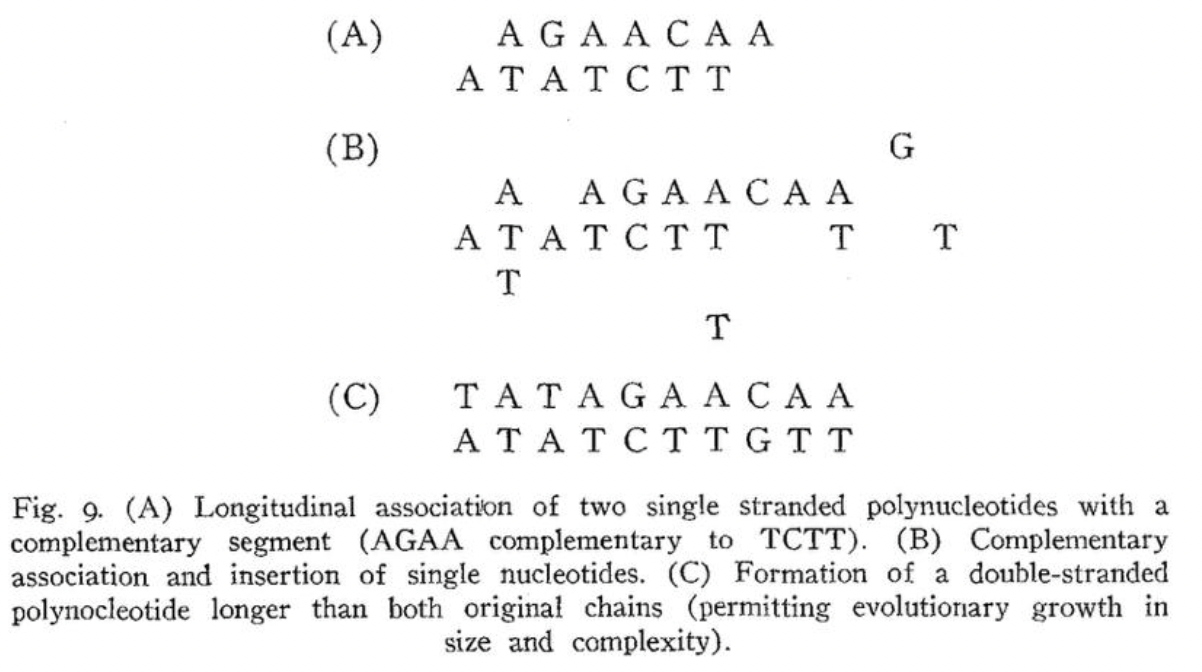}
    \caption{Example of DNA norms proposed by Barricelli.}
    \label{fig:barDNA}
\end{figure}

\begin{figure}[t]
    \centering
    \includegraphics[width=0.4\textwidth]{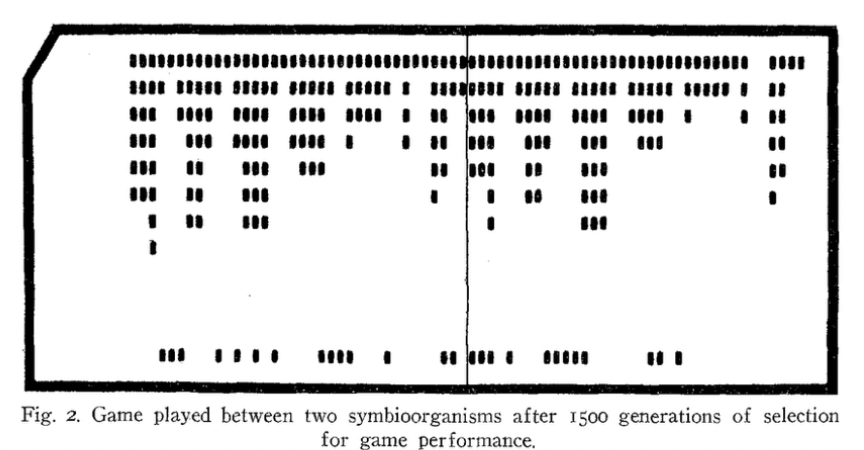}
    \includegraphics[width=0.5\textwidth]{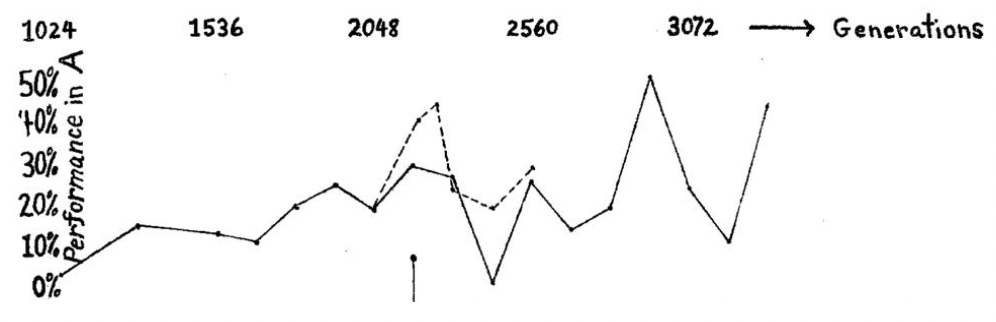}
    \caption{Example of evolutionary computation and embodiement proposed by Barricelli.}
    \label{fig:barevo}
\end{figure}

\begin{itemize}
    \item First known experiments on artificial organisms and simulated evolution \citep{barricelli1954esempi, barricelli1957symbiogenetic, barricelli1962numerical, barricelli1963numerical}
    \item IMPORTANT NOTE: there is evidence today that major evolutionary transitions, such as the one from eukaryotes to prokaryotes, were the result of symbiogenetic events \citep{szathmary1995major}. There is additional evidence by looking at the human genome, where only a small part codes for proteins and a much larger part codes for retroviruses that are integrated in our genome~\cite{Aguera2025WhatIsLife}
    \item“The first sexually reproducing organisms have arisen as a result of a symbiosis between self-reproducing elements (genes), e.g., DNA or protein molecules with the capacity to reproduce by autocatalytic processes and the capacity to undergo certain hereditary changes (mutations).”
    \item”Genes were originally independent organisms which have united by symbiosis to form a kind of primitive chromosome“
    \item“Even though the original genes individually had a very limited reportoire of mutations capable of producing viable varieties, … symbiosis might lead to the formation of organisms with a much larger repertoire of mutations …”
    \item Barricelli’s goals: \\
    1) Support a symbiogenetic theory of the origin of life (Darwinian evolution alone is not sufficient to explain how life started) \\
    2) Observe how a symbiogenetic evolution process starts and initially develops by hereditary changes and selection; is symbiogenetic evolution open-ended?
    \item Artificial organisms are numbers in a 1D finite discrete space (512 cells on the IAS machine due to only 5 Kb of memory), time is discrete (generations), organisms must have rules for reproduction, mutation, and interaction. 0 = empty cell, Non-zero natural number = a gene
    \item \textbf{Multi-gene individuals emerge spontaneously} (see Fig. \ref{fig:bar57organism}) only if elements (numbers) on the grid are allowed to \textbf{replicate} (in his work this is implemented by shift rules moving numbers to a shifted location equivalent to their values, positive to the right and negative to the left), to \textbf{mutate} (when two numbers collide on the same location, a mutated version is produced for example by summing them or killing them both - note that Barricelli proposed different policies called norms), and to \textbf{interact via symbiosis} (multiple replications could occur via interacting numbers, that is when a number lands on a location occupied by another number at the previous generation; in that case, the shifted number would replicate a second time in a location shifted from its original location of a quantity dictated by the interacting number present at the previous generation). See Fig. \ref{fig:barsymbiosis}
    \item Additional findings by Barricelli: 
    \begin{itemize}
    \item emergent sexual recombination via crossover (see Fig. \ref{fig:barcross})
    \item uniform worlds governed by a single norm show symbioorganism emergence but not open-ended evolution
    \item quasi-uniform CA where different rules are applied in different patches of the environment show prolongued dynamics
    \item parasitism emerges where organisms cannot replicate alone but only in the presence of a host organisms (which may survive or die)
    \item self-repairing organisms would emerge (see Fig. \ref{fig:barrepair})
    \end{itemize}
    \item Barricelli suggests that DNA is using symbiogenetic interaction norms (nucleotides have specific association rules) and drafts a possible experiment to show that single nucleotides (an abstract computational system working on an alphabet of four symbols) and computational symbiogenetic norms could produce the emergence of self-replicating double stranded symbioorganisms, see Fig. \ref{fig:barDNA}
    \item Barricelli runs the (likely) first experiment of evolutionary computation (see Fig \ref{fig:barevo}). He writes: “we must give the genes some toy bricks to play with, … some material they may organise and use in the competition among different symbioorganisms”. He runs experiments where two symbioorganisms competing to replicate in the same location would play a game (TacTix, where the strings of numbers of each symbioorganism would encode for playing strategies). The one that won the game would get the right to reproduce. This selective pressure produced organisms that acquired better strategies over evolution (i.e., learning), therefore encoding both self-replicating programs as well as functional programs to play the game. 
    \item \textbf{Code repo for reproducing various Barricelli's experiments (and more!)  \\ https://github.com/JELAshford/symba-alice-2026}
\end{itemize}

\section{Background}
This background will cover aspects related to ALife, such as:
\begin{itemize}
    \item Symbiosis, symbiogenesis, endosymbiosis, mutualism, commensalism, amensalism, competition
    \item History in biology and machine learning (ML) / Artificial Ingtellicence (AI)
    \item Definitions in ML, Symbiotic Organism Search (SOS) algorithms and survey, loss of Barricelli’s original ideas
\end{itemize}

\subsection{Restrictive vs broad definition of symbiosis in biology}
Heinrich Anton de Bary coined the term in 1879, defining symbiosis broadly as “the living together of unlike organisms” @. This original definition explicitly included all types of close, long-term interactions between different species—mutualism, commensalism, and parasitism @. Shortly after de Bary’s definition, some biologists began restricting “symbiosis” to mean only mutualistic relationships where both organisms benefit. This restrictive usage likely stems from everyday language where “symbiosis” implies cooperation. The confusion was compounded by prominent biologists like Hertwig (1906) who reversed his own earlier position and promoted the restrictive view @. By the late 20th century, a “growing consensus in the biological literature” emerged favoring de Bary’s original broad definition @. Modern sources increasingly define symbiosis as “any close and long-term biological interaction between two organisms of different species,” with mutualism as one specific type. Lynn Margulis adopted and promoted the broad definition in their work on endosymbiosis and holobiont theory @. Contemporary textbooks and encyclopedias now predominantly use this framework, though the restrictive usage persists in popular discourse. In the rest of this study, we will use the broad definition: symbiosis as “the living together of unlike organisms” including all beneficial (mutualism, commensalism), neutral and harmful interactions (amensalism, competition). From this perspective, we can visualize a continuum between antagonistic and cooperative symbiotic relationships.

\subsection{Related Works}

\subsubsection{Biological Symbiogenesis Theory}
The biological case for symbiogenesis as a major evolutionary force was made most influentially by Margulis, whose Serial Endosymbiosis Theory (SET) proposed that mitochondria and chloroplasts originated as free-living bacteria that entered into obligate symbiotic relationships with their host cells, driving the emergence of eukaryotic complexity \citep{margulis1970origin}. Margulis adopted de Bary's broad definition of symbiosis (discussed in Section~3.1) and placed it within an evolutionary framework in which symbiogenesis, not merely gradual mutation and selection, is a principal engine of major evolutionary transitions \citep{szathmary1995major}. Her work motivates the computational investigations pursued in this report: if symbiogenesis can drive the emergence of complex life from simpler biological entities, can analogous processes produce emergent complexity in computational substrates?

\subsubsection{Barricelli's Computational Symbiogenesis}
Barricelli's work on numerical symbiogenesis, reviewed in detail in Section~2, is the most direct predecessor of the present study. Across a series of publications from 1954 to 1963 \citep{barricelli1954esempi, barricelli1957symbiogenetic, barricelli1962numerical, barricelli1963numerical}, he showed that simple numerical organisms in a one-dimensional cellular automaton could, through replication, mutation, and symbiotic interaction, give rise to self-reproducing multi-gene structures he called symbioorganisms. Replication and mutation alone produced limited variability, whereas symbiotic association allowed even elements with only two allelic states to form organisms with $2^n$ varieties \citep{barricelli1962numerical}. Fogel gives a historical account positioning Barricelli as a pioneer of artificial life decades before the field was formalized \citep{fogel2006barricelli}. Our work builds on Barricelli's framework: we replicate and extend his 1D systems with quantitative analysis, generalize them to two dimensions, and introduce DNA-inspired norms that he proposed but never implemented.

\subsubsection{Self-Replication in Computational Systems}


Von Neumann introduced self-reproducing cellular automata based on a universal constructor \citep{von_neumann1966theory}, and Langton later showed that self-reproduction could be achieved with far simpler structures using minimal self-reproducing loops \citep{langton1984self}. Both focused on engineering individual replicators. Barricelli's approach differs in that complex self-replicating structures are not designed but emerge from populations of simple interacting elements.


A separate line of work explored self-replication and evolution in assembly-like computational substrates. Pargellis showed evolutionary emergence in the Amoeba system, where programs composed of assembly-like instructions could evolve through variation and selection \citep{pargellis1996evolution}. This tradition, surveyed by Taylor et al.\ from von Neumann through Tierra and Avida to modern systems \citep{taylor2020rise}, shares with Barricelli's numerical organisms the elements of self-replication and open-ended evolution. These systems generally lack symbiogenesis mechanisms, however: organisms evolve through mutation and selection but do not form persistent cooperative associations that give rise to new kinds of entities.


\citet{alakuijala2024computational} showed that well-formed, self-replicating programs can emerge from random Brainfuck family (BFF) programs through interaction alone, without explicit fitness landscapes. Their work introduced high-order entropy as a measure of emergent complexity and a token-tracing methodology for tracking gene ancestry. Interestingly, the parallel to Barricelli's symbioorganisms is quite close: in both systems, self-replicating structures arise not by design but as consequences of local interaction rules. The Computational Life framework operates with a fixed interpreter (the BFF compiler), however, whereas Barricelli's norms define the interaction rules themselves, a distinction we return to in Section~7.

Nearly all of the systems above operate on one-dimensional substrates. Higher-dimensional CAs have been explored elsewhere, particularly in the Game of Life tradition and in Lenia \citep{chan2019lenia}, but none of these incorporate the shift-based replication and collision norms central to Barricelli's framework. Barricelli himself speculated that similarly complex dynamics might emerge in two-dimensional variants of his automaton, but to the best of our knowledge he never implemented or published such experiments. Our 2D extension (Section~5.2) addresses this gap.

\subsubsection{Symbiogenesis in Evolutionary Theory and Computation}


Several lines of theoretical work have sought to formalize the conditions under which symbiogenesis can drive evolutionary transitions. \citet{bull1995artificial} modeled when endosymbiosis becomes advantageous. Watson and Pollack showed that symbiotic composition enables evolutionary transitions by allowing hierarchical organization: entities that could not solve complex problems individually can do so when composed symbiotically \citep{watson2001symbiotic, watson2003computational}. Daida extended symbiogenesis to complex adaptive systems more broadly \citep{daida1996symbionticism}, while \citet{egbert2023behaviour} argued that self-organization and interaction patterns, which are central to Barricelli's requirement that symbiotic cooperation precede complex evolution, are precursors to biological organization itself. We find that these contributions bear on the phenomena we observe in our own replications, where cooperative structures emerge and persist as replicating units.


Within evolutionary computation, several algorithms have adopted the label ``symbiotic,'' though most use the term loosely. The Symbiotic Evolutionary Algorithm of \citet{halavati2007general, halavati2009symbiotic} addresses linkage problems by allowing cooperative relationships between gene groups, inspired by Barricelli's observation that symbiosis enables complex trait evolution. It treats symbiosis as a heuristic for gene grouping, rather than as an emergent dependency between interacting entities. Symbiogenesis has also been explored as a mechanism for building complex adaptive systems in genetic programming \citep{heywood2010symbiogenesis}, with benchmarking studies validating the scalability of symbiotic coevolutionary approaches \citep{doucette2012symbiotic}. What these methods share, in our view, is that they impose symbiotic structure as an algorithmic design choice rather than allowing it to emerge from the dynamics of the system, as it does in Barricelli's original framework and in the experiments we present here.

\subsubsection{Biochemical and Computational Analogies}

\citet{akhlaghpour2022rna} argued that RNA can serve as the basis for natural universal computation through self-replication and catalysis, identifying combinator operators in base-pairing and splicing that render RNA Turing-complete. Holland's work on genetic operators \citep{holland1992adaptation} formalizes variation and recombination in abstract computational terms. Both of these perspectives are relevant to the DNA-norm expansion we introduce in Section~5. Barricelli proposed extending his numerical norms toward a more explicitly DNA-like regime in his later writings but, as far as we are aware, never implemented them. Our minimal DNA-norms (elongation, complementary association, and strand separation in a nucleotide soup) are a first step toward connecting Barricelli's abstract numerical symbiogenesis with the biochemistry of replication.

\section{Extending Views on Symbiosis}

\subsection{Symbiosis, a multiscale phenomenon in biology and evolution}

Many parasites and pathogens fill the definition of symbiosis. We can include low-virulence ones like Helicobacter pylori, known to infect the stomach of mammals and discovered as a driver of stomach cancer, Chlamydia bacteria, or herpes viruses.

Another emblematic case of symbiosis is endosymbiosis, more specifically the process by which an organism is integrated in another for the benefits of both species. Symbiosis drove the evolution of eukaryotic complexity—mitochondria and chloroplasts originated as free-living bacteria that became obligate symbionts.



\textbf{An important question is whether symbiogenesis should be understood purely as an evolutionary phenomenon or whether analogous processes occur in non-evolutionary contexts such as development, physiology, or social organization. 
}

Classical accounts treat symbiogenesis as a macroevolutionary event in which previously independent replicators become permanently integrated into a new reproductive unit, as in the incorporation of mitochondria into eukaryotic cells \cite{margulis1981symbiosis,margulis1971symbiosis,smith1997major}. In this view, symbiogenesis produces a new evolutionary individual whose components can no longer reproduce independently. However, similar irreversible integrations may occur within the lifetime of an organism. Developmental processes often stabilize cooperative cellular collectives that cannot revert to their earlier autonomous states, for example when cells differentiate and become functionally integrated into tissues or organs. Likewise, symbiotic interactions with microbiota or other organisms can become developmentally entrenched, forming tightly coupled functional systems sometimes described as holobionts \cite{gilbert2012symbiotic}. 

These cases suggest that symbiogenesis may be understood more broadly as a dynamical process in which initially independent agents become progressively integrated through feedback, coordination, and constraint until they form a new level of organization. Similar transitions appear in collective systems such as swarms or social groups, where coordinated interactions can produce higher-level entities with emergent regulatory properties~\cite{okasha2006evolution,west2015major}. In human societies, organizational structures--from institutions to scientific communities--also display forms of functional integration in which the collective acquires capabilities not reducible to individual participants. From a connectionist perspective, such transitions can be interpreted as the emergence of new higher-level ``modules'' in which interactions among components become structured and canalized, enabling the collective to function as an adaptive unit and allowing evolving and, more generally, learning systems to exploit previously learned patterns of coordination~\cite{watson2016evolutionary,watson2016can,watson2022design,kouvaris2017evolution}. These observations motivate a dynamical view of individuality, in which individuals are not static objects but metastable patterns of interaction maintained through ongoing processes of coordination and constraint. From this perspective, individuality emerges when a collective achieves sufficient integration, causal closure, and alignment of function to behave as a coherent agent over time \cite{godfrey2015individuality}. The transition from symbiosis to symbiogenesis can thus be interpreted as a key mechanism by which collectives become individuals.

\subsection{Symbiosis as a process of collective intelligence}

By using the extended definition of symbiosis, we can find parallels between collective intelligence processes and symbiosis (mutualism, parasisitsm, commensalism).



\textbf{Mutualism =  cooperative behavior = scaling of goals/intelligence/survival}

By sharing stress, cells can resolve more complex problems, allowing the scaling of goals \citep{Levin2019BoundaryOfSelf, PioLopez2023ScalingOfGoals}. Game-theoretic results shows also that when agents decide to merge, they increase their problem-solving capacities, see the game-theoretic results on agents and superagents \cite{Shreesha2025}.

\textbf{Commensalism = neutral behavior (or that seems neutral)}

Studies have shown that unilateral relationships—such as commensalism and amensalism— can strengthen community stability more effectively than symmetric interactions like competition or mutualism \citep{mougi2016roles}. Ecological communities that combine different types of unilateral interactions tend to be more resilient, indicating that asymmetric relationships, where contributions are uneven, can foster robust collective behavior.

\textbf{Parasitism = Competition (that can be beneficial too)}

Collective intelligence is a process of cooperative/competition as multiple scales \citep{watson2023collective}. Competition can be beneficial when we jump scale. Similarly to why parasitic behavior can be beneficial by reducing competitive pressure and increasing community biodiversity, for  example~\citep{hatcher2012diverse, gomez2013neglected, hudson2006healthy}.

Intelligent behavior is proposed to arise from balancing cooperative and competitive dynamics at the same time. Competition within a team can enhance collective intelligence when it boosts integrative information sharing while simultaneously increasing uncertainty for rival teams. Competition between teams imposes selection pressures that favor better coordinated groups, even in the absence of explicit coordination mechanisms \citep{hristovski2020theory}.

\section{Implementation of Barricelli Systems}

Barricelli conducted several experiments on symbiogenesis using a simple numerical system, however, his results were limited by the computational resources he had. We re-implemented the system he used in \citep{barricelli1957symbiogenetic}. The code repo for reproducing various Barricelli's experiments (and more!) is available at the following link: \\ https://github.com/JELAshford/symba-alice-2026

\subsection{1D Barricelli System}

\begin{figure}[ht]
    \centering
    \includegraphics[width=0.40\textwidth]{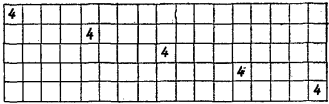}
    \\
    \includegraphics[width=0.50\textwidth]{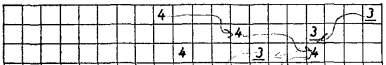}
    \caption{Barricelli's numerical system. On the top, an example of a single number replication. In the bottom, the number 4 is replicated twice, due to interaction with the number -3.}
    \label{fig:barricelli-system}
\end{figure}

Barricelli's original system consisted of 1D arrays of integers stacked into a 2D array $X$, where $X_{a, g}$  represents the value of the array at position $a$ in generation $g$. To compute the next generation, the algorithm does as follows:

1. A number $n$ at position $a$ in the 1D array attempts to replicate by copying itself into position $a+n$ (Figure \ref{fig:barricelli-system}, top). If there is already a non-zero value $m$ in that position, the $n$ also attempts to copy into $a+m$ (Figure \ref{fig:barricelli-system}, bottom). This process repeats until $n$ copies into a previously unoccupied location, or attempts to copy into the same location twice.

2. If multiple numbers try to replicate into the same position, a norm (also referred to as a `mutation rule`) is used to resolve the collision. The norms that Barricelli used are described in Table \ref{tab:collision-norms}).

\begin{table}[ht]
\caption{Barricelli's collision norms.}
\label{tab:collision-norms}
\centering
\small
\begin{tabularx}{\textwidth}{|c|X|X|}
\hline
\textbf{Name of the norm} &
\textbf{$X_{ag}$-value (computed by mutation or exclusion)} &
\textbf{Conditions} \\
\hline
Norm 0 & $0$ & Collision of different genes. \\
\hline
\multirow{3}{*}{Norm A}
& $u + v$ & Collision under empty cell if $X_{a+u,\, g-1}$ and $X_{a-v,\, g-1}$ have same sign. \\
\cline{2-3}
& $-(u + v)$ & Same collision if the sign is not the same. \\
\cline{2-3}
& $0$ & Collision under occupied cell. \\
\hline
\multirow{3}{*}{Norm B}
& $u + v - 1$ & \multirow{3}{\hsize}{Respectively same conditions as in A norms.} \\
\cline{2-2}
& $-(u + v - 1)$ & \\
\cline{2-2}
& $0$ & \\
\hline
\multirow{2}{*}{Norm C}
& $X_{a-v,\, g-1} - X_{a+u,\, g-1}$ & Collision under empty cell. \\
\cline{2-3}
& $0$ & Collision under occupied cell. \\
\hline
\multirow{2}{*}{Norm D}
& $-X_{a,\, g-1} + 2X_{(a + X_{a,\, g-1}),\, g-1}$ & Collision if $X_{(a + X_{a,\, g-1}),\, g-1}$ and $X_{(a - X_{a,\, g+1}),\, g-1}$ are equal. \\
\cline{2-3}
& $0$ & Same collision if the two numbers are not equal. \\
\hline
\end{tabularx}
\end{table}

\begin{figure}[ht]
    \centering
    \includegraphics[width=0.40\textwidth]{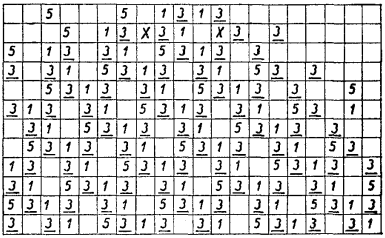}
    \caption{Barricelli's symbioorganism.}
    \label{fig:symbioorganism}
\end{figure}

Barricelli reasoned about self-replicating sequences of numbers as symbioorganisms. One of the examples he provides is shown in \ref{fig:symbioorganism}. In \citep{barricelli1962numerical}, he claims these patterns share many properties with living organisms, such as self-replication, mutations, parasitism, robustness to damage, and more. However, Barricelli lacked resources to verify and present these properties rigorously.

He also conducted several \textit{large scale} experiments, up to 512 cells and 5000 generations (See Figure. He describes the outcome of these simulations as "organized homogeneity", i.e. a single "simple symbioorganism" would take over, and the universe would enter periodic behavior. Barricelli claimed that several augmentations successfully solved this issue, including using multiple norms on the same grid and launching parallel simulations that periodically exchange their segments \citep{barricelli1957symbiogenetic,  barricelli1962numerical}.

\begin{figure}[t]
    \centering
    \includegraphics[width=0.29\textwidth]{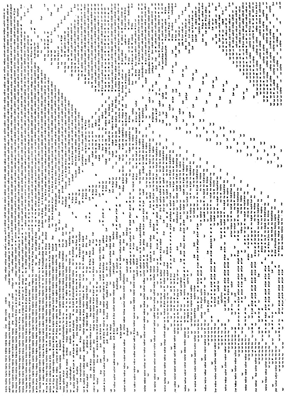}\hspace{0.01\textwidth}
    \includegraphics[width=0.65\textwidth]{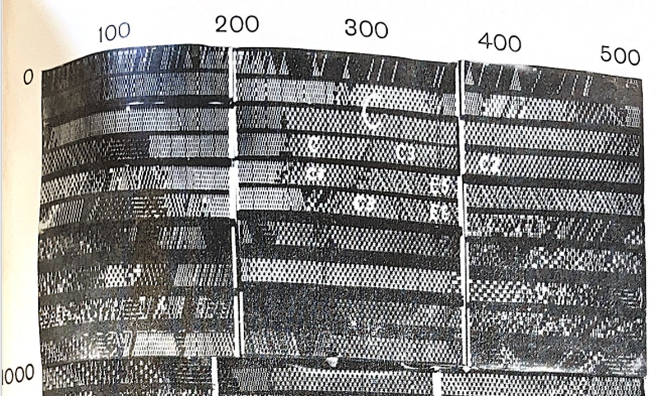}\hspace{0.01\textwidth}
    
    \caption{Barricelli's reports on \textit{large scale} experiments. \textbf{Left:} Barricelli system initialized with random numbers converges towards symbioorganisms. Reproduced from Figure 15 of \citep{barricelli1962numerical}. \textbf{Right:} Barricelli system with heterogeneous norms, annotations demarcate different norms and symbioorganisms. Reproduced from Figure 9 of \citep{barricelli1957symbiogenetic}.}
    \label{fig:barricelli_large_scale}
\end{figure}

\begin{figure}[!ht]
    \centering
    \includegraphics[width=0.32\textwidth]{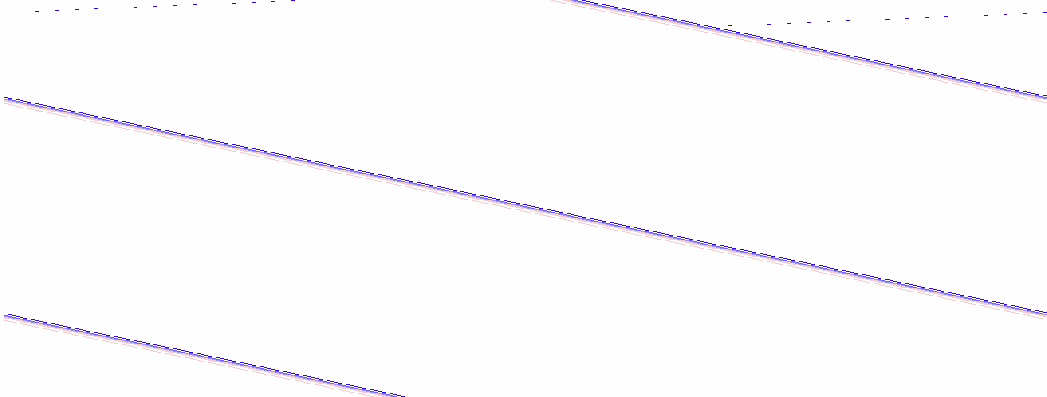}\hspace{0.01\textwidth}
    \includegraphics[width=0.32\textwidth]{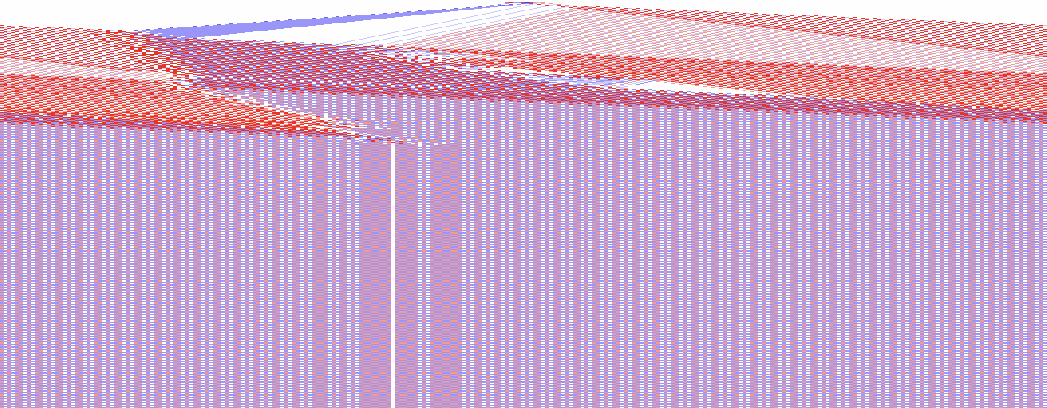}\hspace{0.01\textwidth}
    \includegraphics[width=0.32\textwidth]{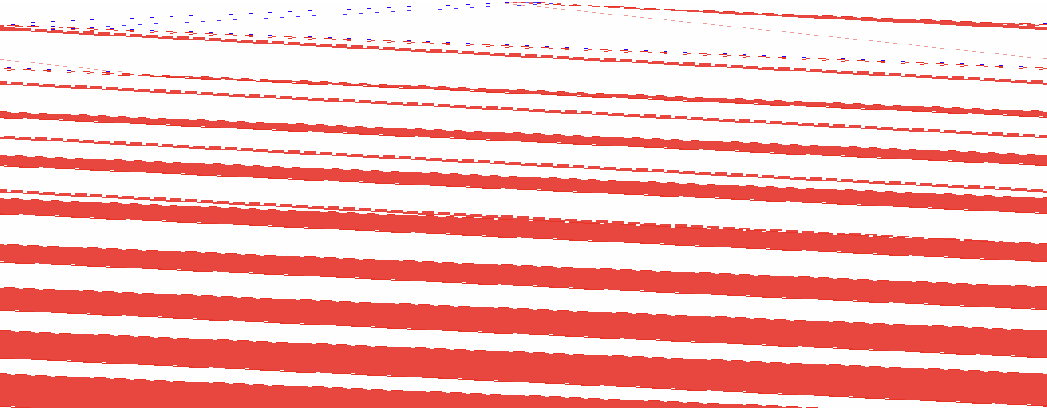}\\[0.5em]
    \includegraphics[width=0.32\textwidth]{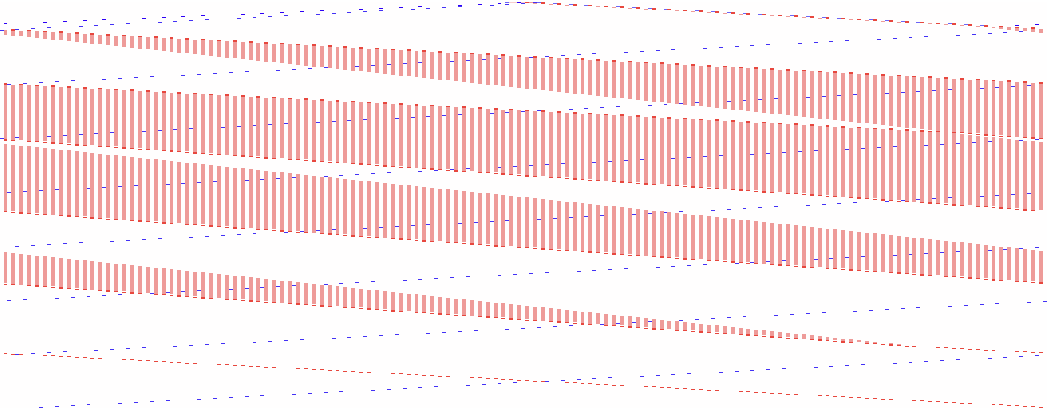}\hspace{0.01\textwidth}
    \includegraphics[width=0.32\textwidth]{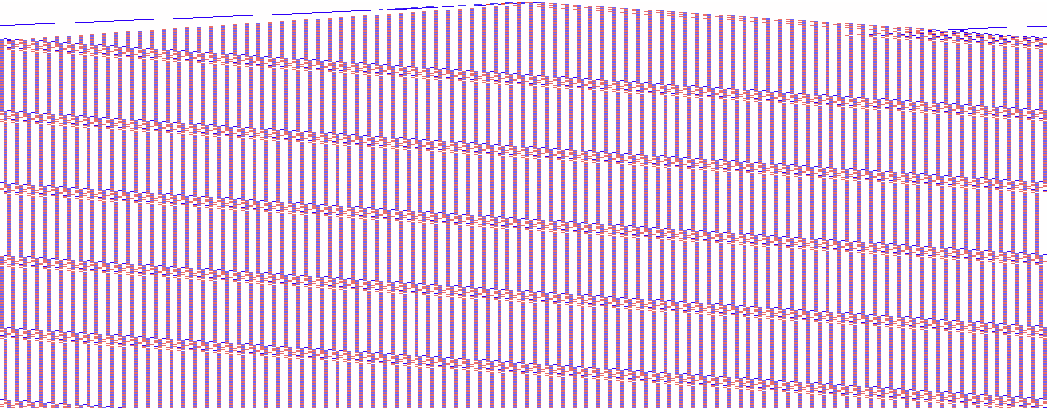}\hspace{0.01\textwidth}
    \includegraphics[width=0.32\textwidth]{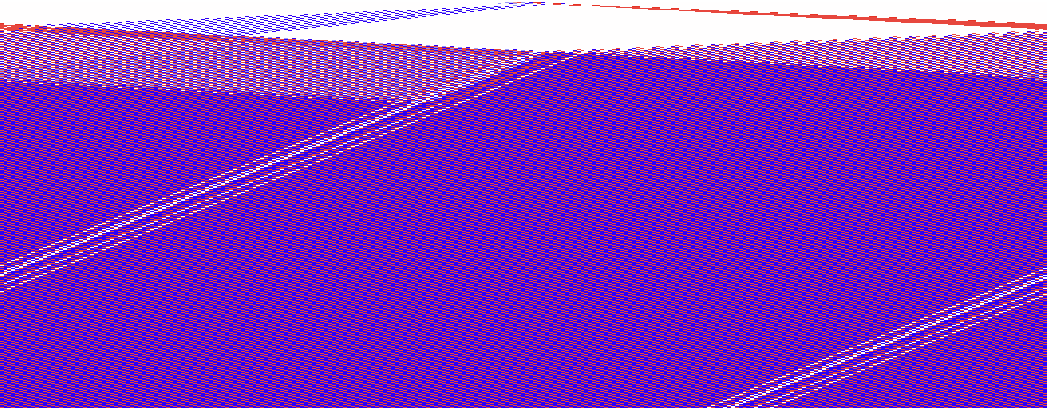}
    \caption{6 examples of Barricelli's numerical systems with $0$ norm and periodic boundary conditions. The grid of length 256 is initialized with 10 random numbers in the center, the rest are 0. The simulations ran for 512 generations.}
    \label{fig:norms}
\end{figure}

\subsection{2D Barricelli System}
Barricelli makes reference to 2D variants of this automata in multiple papers, speculating that similarly complicated dynamics that are seen in the 1D worlds might be exhibited there also. \textbf{It appears that due to hardware or time constraints he was not able to share any examples of this behavior, so we have implemented it ourselves here}. 

Our implementation follows an identical conceptual structure to the 1D case, with the extension that instead of 1D array of integers for each timestep there is a 2D array of integer pairs. These pairs describe a direction to move during replication, which is carried out exactly as in the 1D case, and then collisions are resolved with norms. 

In our experiments, the $0$ and $D$ norms can be used unmodified, as they consider either the presence or absence of a collision, or a step in the cell's current "direction" respectively - both of which are independent of the dimensionality of the substrate. 

As in the 1D cases, we observe that originally sparse starting conditions seed with positive and negative numbers quickly organizes into denser fields with repeating and self-propagating patterns.

\begin{figure}[!ht]
    \centering
    \includegraphics[width=1\textwidth]{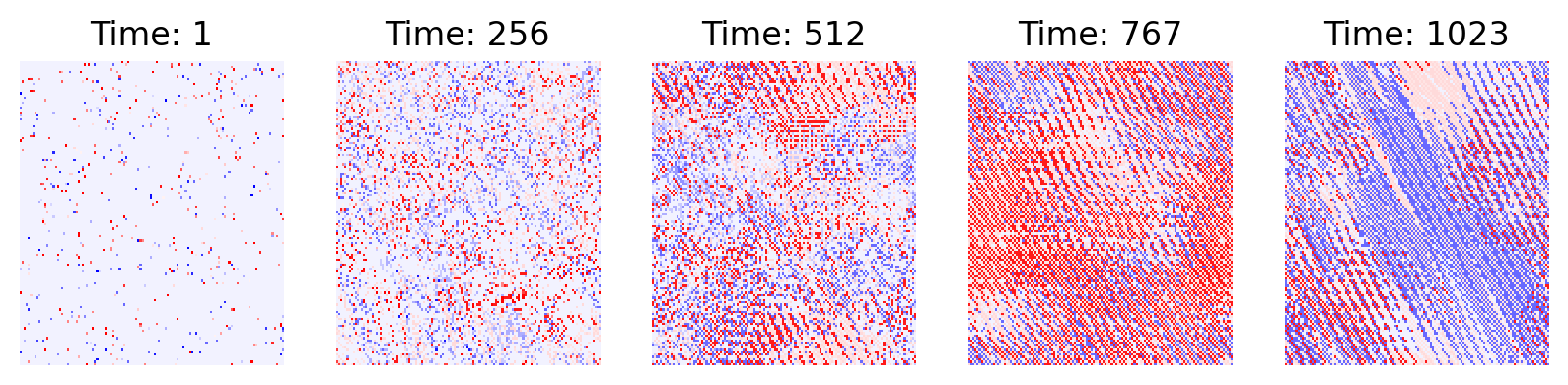}
    \caption{Barricelli's $0$ norm executed on a $2D$ environment, with an 80\% sparse initialization and toroidal wrapping. Each point represent the $2D$ vector at that position by calculating it's polar angle with $atan2$ and applying a divergent color scheme. \textbf{A video of the full run is available \href{https://github.com/JELAshford/symba-alice-2026/blob/757a34b38f547400e9664a6ddb4490facc2c4071/out/2d.mp4}{here}}.}
    \label{fig:2d_barricelli}
\end{figure}

\subsection{DNA-norm Expansion}
\label{sub:DNA:norm}

Barricelli's numerical symbiogenesis experiments showed that, even in a highly abstract one-dimensional medium, persistent patterns can reproduce, compete, and combine into higher-order "symbioorganisms". A central methodological point in that program is that the outcomes depend critically on the norms (i.e., the local laws that govern copying, variation, and interaction). 

In later discussions, Barricelli suggested expanding these laws toward a more explicitly DNA-like regime. Motivated by that direction, we expand the symbiogenetic setting by replacing pattern-level replication rules with a minimal set of DNA-norms. We study these norms first in a well-mixed "soup" of nucleotides (monomers) and polymers (single strand), with transient bound complexes; then we implement them in a spatially explicit 1D CA. The goal is to preserve Barricelli's  emphasis on emergence from local interactions while matching the mechanisms interpretable in molecular terms.

\subsubsection{Minimal DNA-norms}

Our implementation represents the environment as a pool of monomers {A, C, G, T} and single-stranded polymers and double-stranded complexes. 
To make polymer growth and association well-defined, strands are treated as oriented sequences with a fixed polarity (a 5' and 3' end).
Complementarity is defined by an involution:

\begin{equation}
\mu:\{A,C,G,T\}\rightarrow\{A,C,G,T\}, \qquad \mu(A)=T,\ \mu(T)=A,\ \mu(C)=G,\ \mu(G)=C,
\end{equation}

\begin{figure}[H]
    \centering
    \includegraphics[width=0.7\textwidth]{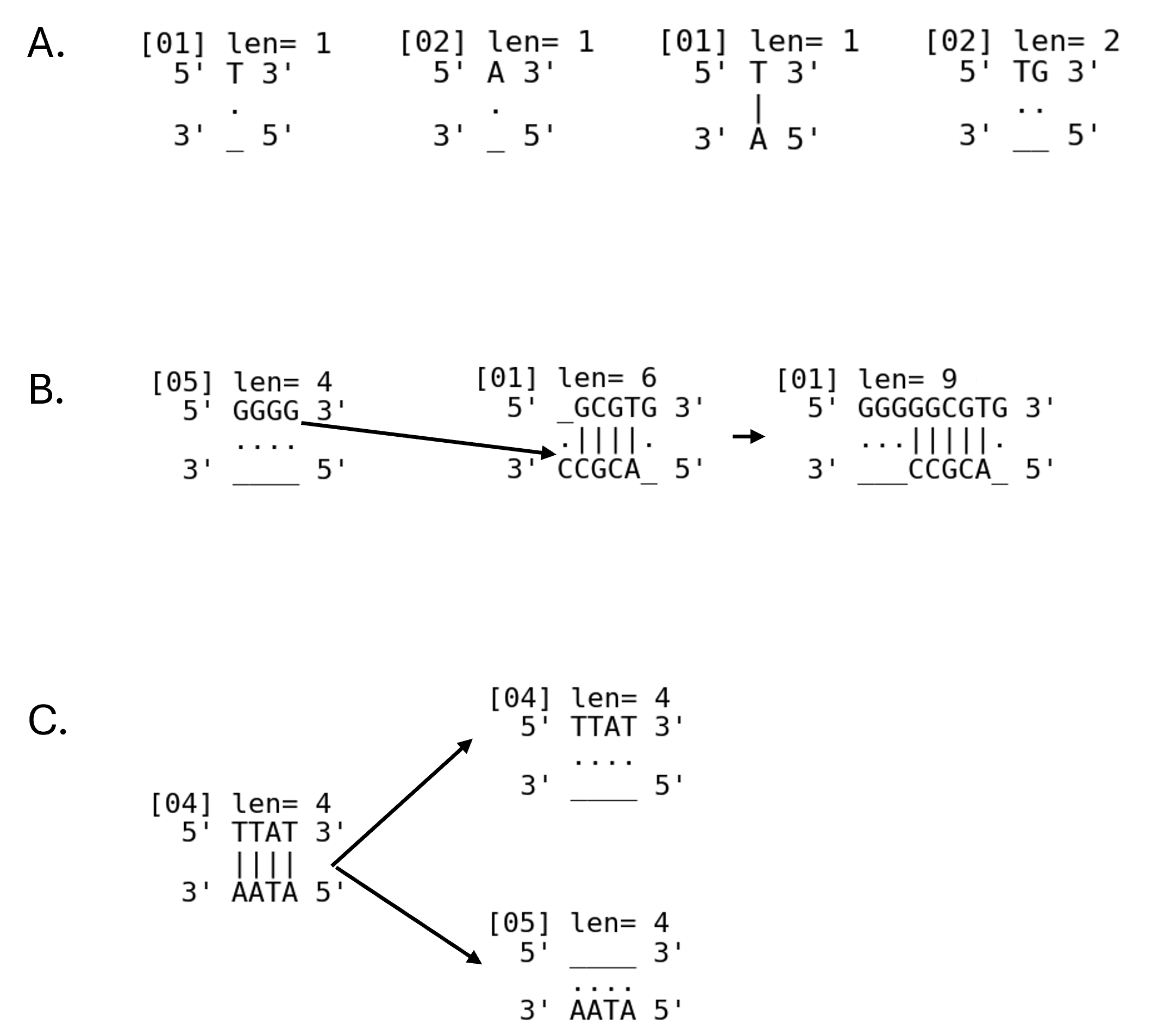}
    \caption{DNA norms used in our implementation.}
    \label{fig:dna-rules}
\end{figure}

This mapping defines which monomers can pair at a given template position. From here, we implement three norms:

\begin{enumerate}
    \item \textbf{Elongation (Monomer addition and polymer growth).}
    Polymers grow by sequential addition of monomers at the 3' end, increasing strand length one nucleotide at a time. At each update a monomer is drawn from the pool (Fig. \ref{fig:dna-rules}A).

    \item \textbf{Complementary association (pairing and overlap-mediated merging).}
     An incoming monomer may bind opposite an unpaired template base (pairing) within an existing single strand and filling gaps. The same pairing rule allows two strands to anneal when they share a sufficiently long contiguous complementary sequence; such annealing can align partial overlaps so that two fragments merge into a longer double strand than either fragment alone (Fig.~\ref{fig:dna-rules}B).
     (Fig. \ref{fig:dna-rules}B). 

    \item \textbf{Split (strand separation and replication).}
    When a double strand contains a sufficiently long sequence, and the gaps are filled via complementary pairing, the double strand (or "parent") can dissociate into two single strands (or "offspring") that re-enter the soup and can participate in subsequent elongation and complementary association. In this way, strand separation implements reproduction: one double strand parent yields two single-stranded progeny (Fig. \ref{fig:dna-rules}C). 
\end{enumerate}

\subsubsection{Experimental conditions}
Before introducing additional layers of organization, we need to first quantitatively evaluate whether the DNA-norm mechanisms measurably increase recurrent sequence motifs in a well-mixed (spaceless) setting. 
We compare two mechanisms:

\begin{enumerate}
    \item \text{Condition A (elongation-only; baseline).}
    Only end-extension is enabled. In particular, we disable complementary association and overlap-mediated completion.
    \item \text{Condition B (full DNA norms: elongation + complementary association + split).}
    Elongation, complementary association with overlap-mediated merging, and splitting are all enabled.
\end{enumerate}

Each condition is simulated for $N=50$ independent seeds over 400 cycles with mutation rate $\mu=10^{-4}$ per nucleotide per cycle.

\subsubsection{Spatial (CA) embedding of the DNA-norm conditions}
To test whether the same norms produce spatial structure when interactions are local, we embedded Conditions~A and~B in a 1D CA. The CA consists of $N=512$ discrete sites arranged with periodic boundary conditions (i.e., a ring/1D torus where site $0$ and site $N-1$ are neighbors) and is evolved for $T=512$ cycles. Each cell holds at most one DNA fragment, which may be a single strand, partially or fully paired double strand. Interactions - elongation, complementary association, and splitting - are restricted to immediate neighbors. Each cell carries a nucleotide "budget" that diffuses to  neighbors and is depleted by growth, so that resources become spatially patchy over time. For each condition and motif lengths $k \in \{4, 6, 8\}$ we constructed spacetime diagrams in which each cell is coloured by the identity of its dominant $k$-mer at every cycle (Fig.~\ref{fig:ca-spacetime}).

\section{Results}
\subsection{Barricelli System Analysis}

\begin{figure}[H]
    \centering
    \includegraphics[width=1.0\textwidth]{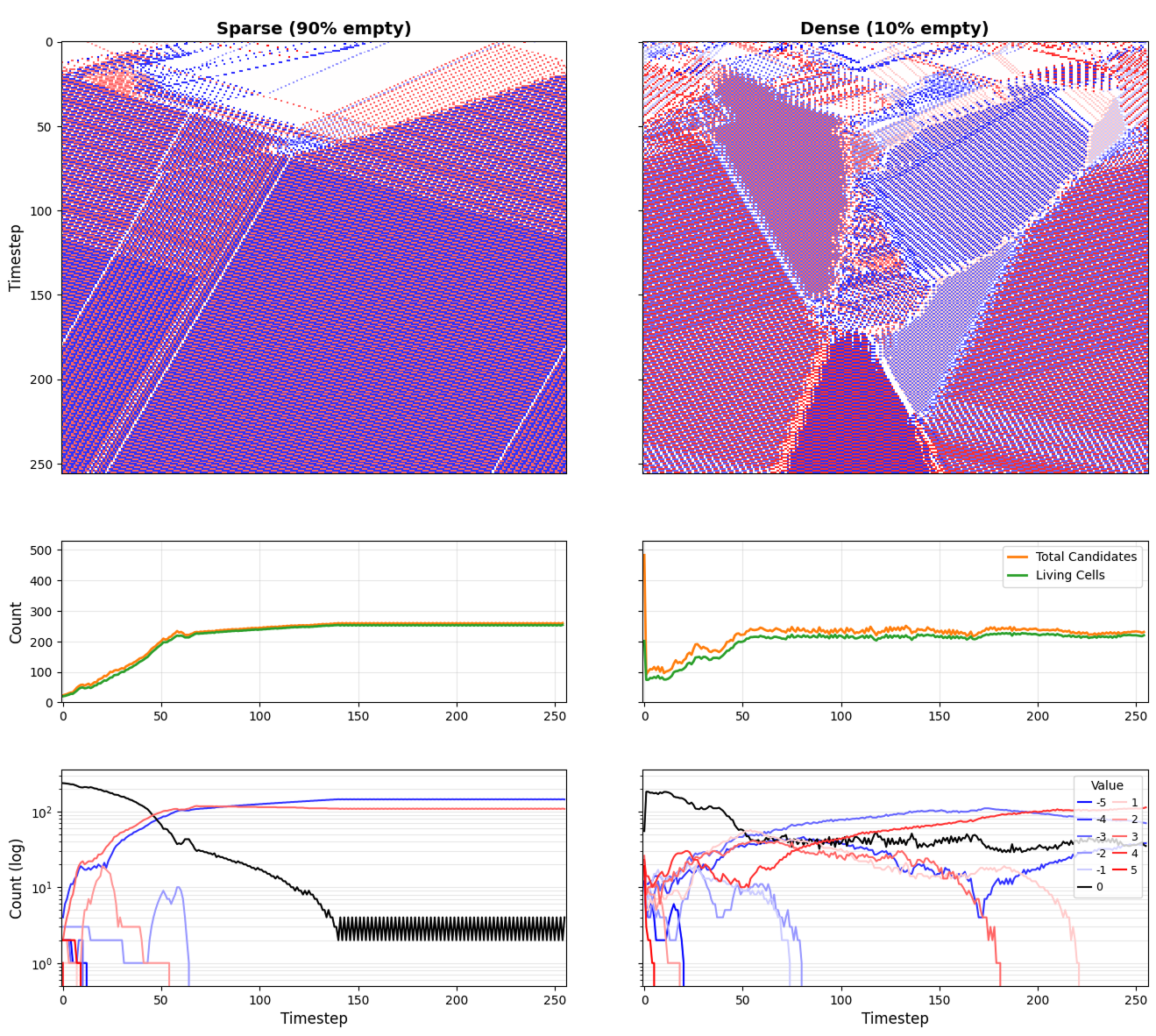}
    \caption{%
        Simple metrics computed for sparse and dense initial conditions over 256 timesteps. The experiments use periodic boundary conditions and $0$ norm.
        \textbf{Top row:} Spatiotemporal evolution of the 1D world.
        \textbf{Middle row:} Aggregate population dynamics showing total replication candidates and living (non-zero) cells over time. Dense initial conditions exhibit rapid population collapse following regrowth, while sparse conditions show gradual population growth as organisms spread into empty space.
        \textbf{Bottom row:} Log-scale distribution of cell values over time, revealing which (numerical values) dominate.
        Under $0$ norm, organisms attempting to replicate into occupied cells mutually annihilate, creating strong selection pressure favouring spatial strategies that minimise conflict.}
    \label{fig:simple_quantiative_measures}
\end{figure}

As mentioned, Barricelli's work lacked a rigorous analysis of the dynamics in the numerical simulations he presents. Many simple metrics could be used to quantify the behavior of the system. For example, one could track population dynamics by counting organisms and collisions over time (as shown in Figure \ref{fig:simple_quantiative_measures}, or their rate of change (i.e., tracking births and deaths). Metrics that track stable numerical sequences would be especially valuable for reasoning about Barricelli's concept of symbioorganisms, though these were not investigated during the workshop.

\subsubsection{Information-theoretic Metrics}
Following work carried out on elementary cellular automata (ECAs), we calculate the individual 1D state entropy over time and investigate the mutual information between states at all time-steps. This allows us to quantify the changes in information over time, without prior knowledge of the system's behavior. 

\begin{figure}[H]
    \centering
    \includegraphics[width=1\textwidth]{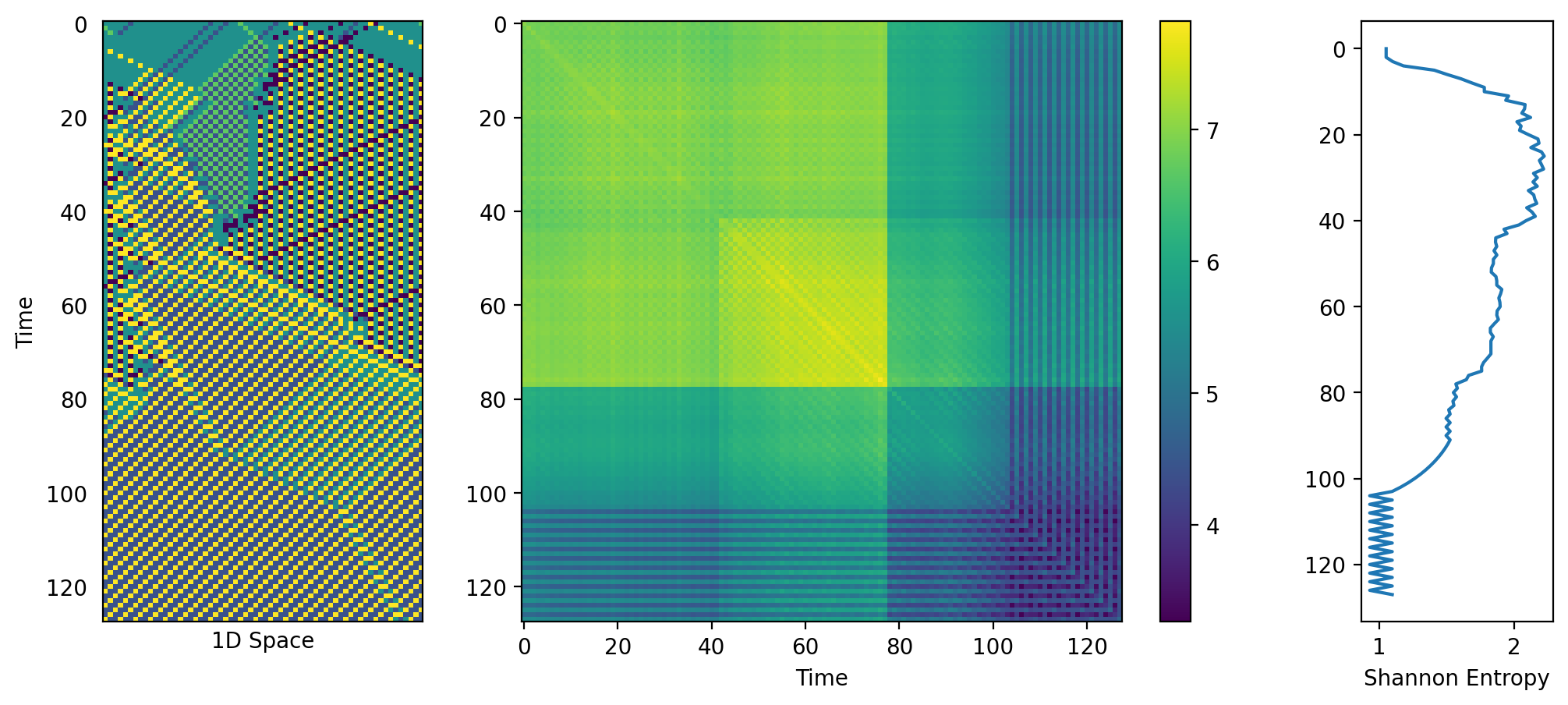}
    \label{fig:1d_information}
    \caption{Information metrics for a $1D$ Barricelli automata over time. The `mutual information` is calculated pairwise between all time-steps, and the `Shannon entropy` is calculated for individual time-steps.}
\end{figure}

In future, we would like to explore the concepts of transfer entropy and locations of active information storage in these CAs, and investigate how these metrics can be efficiently scaled to the $2D$ cases.

\subsection{Robustness of Barricelli's symbioorganisms}

\begin{figure}[H]
    \centering
    \includegraphics[width=0.48\textwidth]{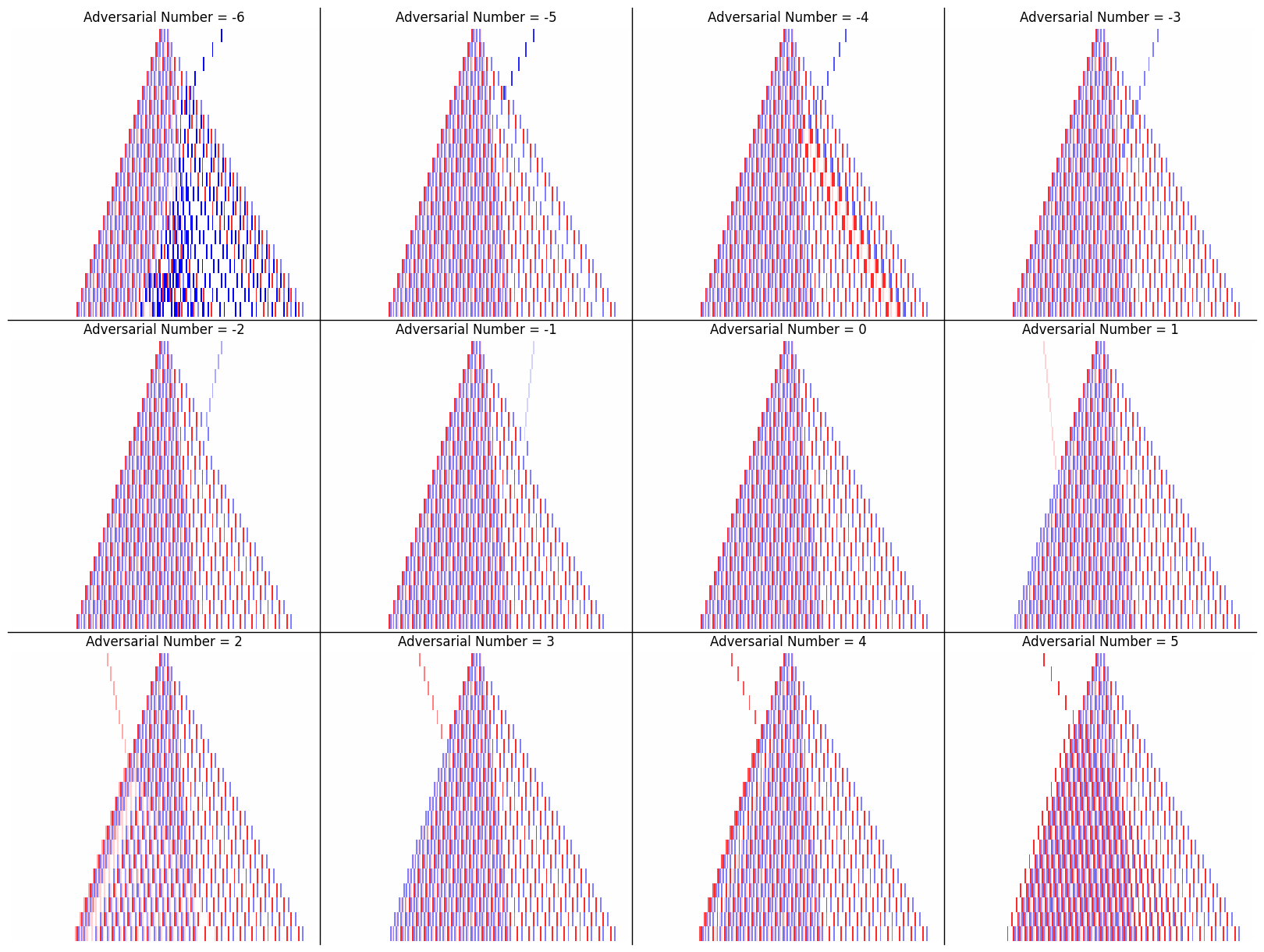}
    \includegraphics[width=0.48\textwidth]{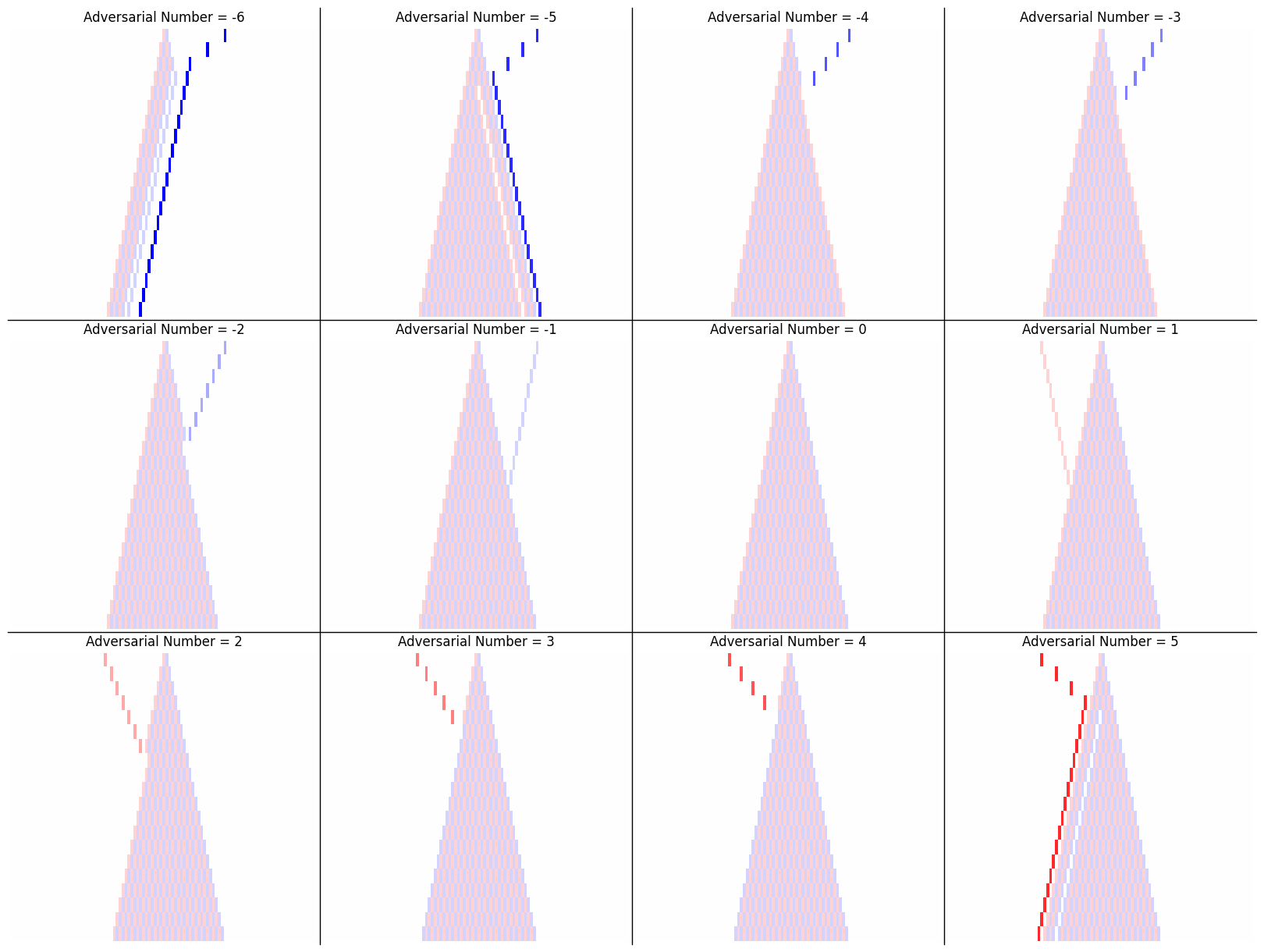}
    \caption{A visualization of how one might study the robustness of Barricelli's symbioorganisms. The grids are initialised with the symbioorganism in the centre. A single additional cell is initialised with a non-zero number that eventually collides with the symbioorganism.}
    \label{fig:robustness}
\end{figure}

Barricelli observed that discovered symbioorganisms exhibit robustness to environmental perturbations and possess the capacity for self-repair, further asserting that "repairing is more likely to succeed the greater the number of symbioorganisms cooperating" \citep{barricelli1962numerical}. Here, we briefly outline a proposed experiment through which these properties may be investigated more systematically. One approach would be to initialise the environment with a symbioorganism and subsequently introduce an additional single numerical value that, as the system evolves, inevitably collides with the symbioorganism. The survival rate of the symbioorganism can then be quantified across a range of conditions (e.g., by varying the introduced numerical values and their initial distances from the symbioorganism) yielding a measure of its robustness (see Figure \ref{fig:robustness}). A complementary line of inquiry would be to examine the symbioorganism's resilience when stochastic noise is introduced into the step function, for example by introducing stochasticity to the norms.

\subsection{2D Cellular Automata Analysis}

A qualitative analysis of some examples in the 2D setup show \textit{interesting} ongoing dynamics and waves propagation.

\subsection{Boolean CA symbiogenesis}

\begin{figure}[H]
    \centering
    \includegraphics[width=0.65\textwidth]{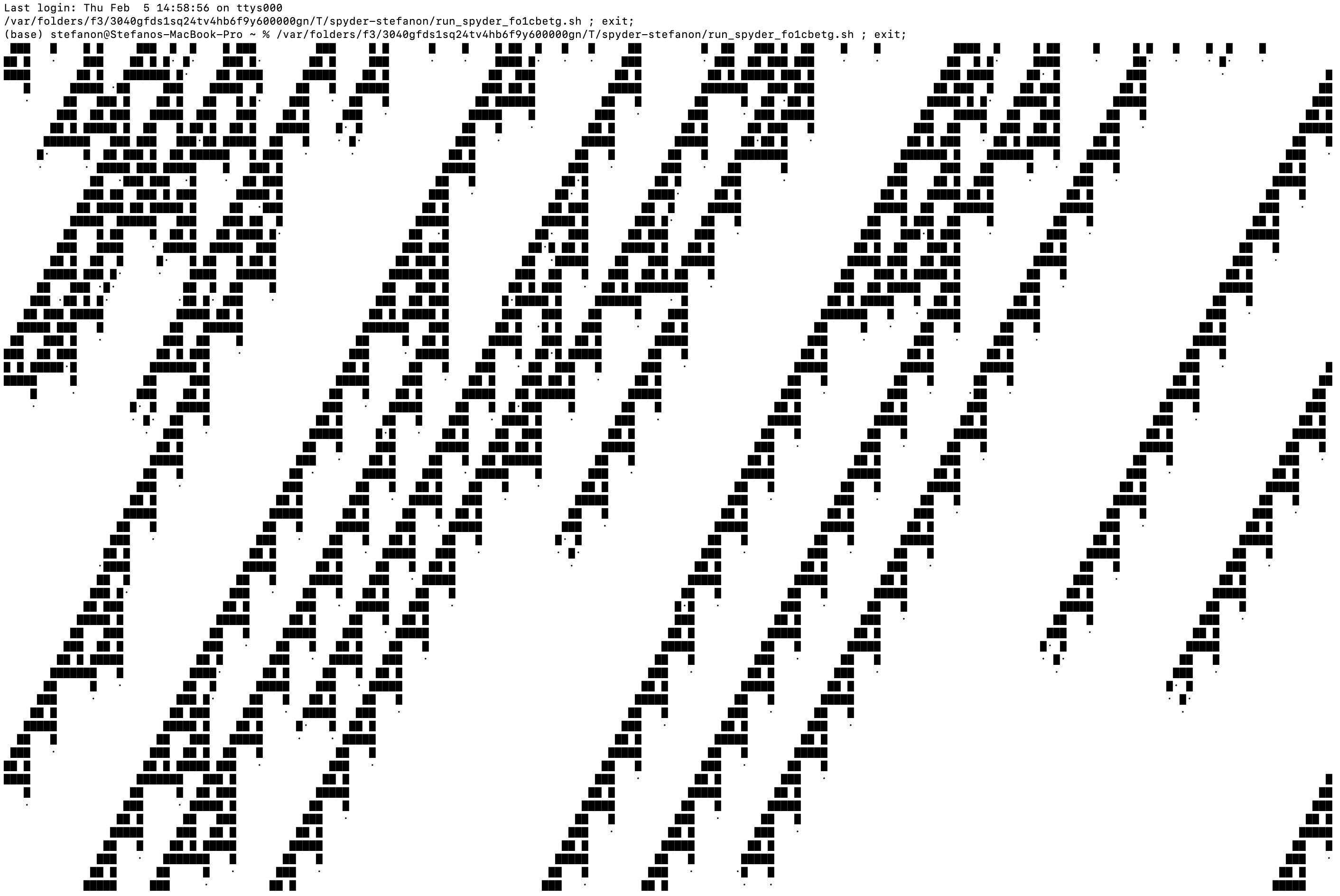}
    \includegraphics[width=0.25\textwidth]{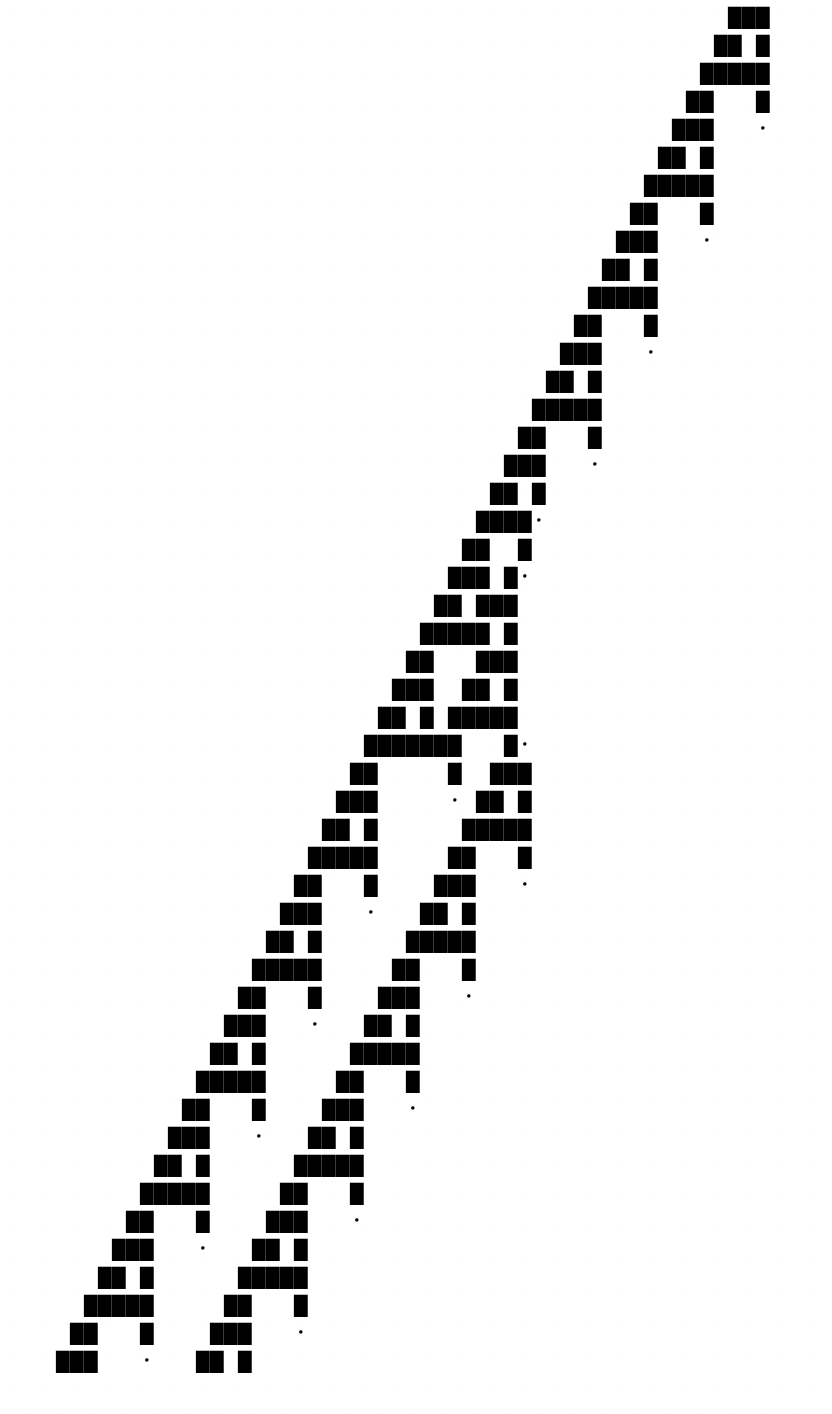}
    \caption{An example Boolean symbiogenesis. Left: from random initial state; Right: detailed view.}
    \label{fig:booleansymb}
\end{figure}

What is the simplest substrate where symbiogenesis effects can be shown? We attempt at experimenting with a 1D Boolean CA where update rules (such as rule 110) would be gated by a symbiogenetic norm (for example whether in the larger neighborhood , that is 2 cells on the left and on the right, for a total of 5 cells including the cell itself - at least a certain number of cells should be active). In case the symbiogenetic norm does not activate the update rule, a decay (mutation) operator is applied where active cells become inactive. Figure \ref{fig:booleansymb} shows an execution of a gated 110 rule, where dots reporesent decay mutations. On the right, a zoomed representation of a branch where a persistent pattern (a Boolean symbioorganism?) duplicates.

\subsection {Preliminary DNA-norm expansion of Barricelli’s symbiogenesis program}
\label{sub:DNA:results}

We compared a baseline elongation-only condition (Condition A) to a DNA-norm condition (Condition B) that additionally permits complementary association with overlap-mediated merging and reproduction by splitting.

Our goal is to quantify when the population contains enough repeated motifs of length $k$ that repetition is no longer a rare accident of sampling. We operationalize this using a repetition threshold $\tau$. At each cycle $t$, we collect all contiguous substrings of length $k$ ("$k$-mers") from all strands in the soup. This produces a multiset of $k$-mer windows (sliding windows along each strand).
From these windows we compute the repeated-window fraction $R_k(t)$ : the fraction of windows whose $k$-mer occurs at least twice in the population in that cycle. If $W_k(t)$ is the total number of extracted windows and $S_k(t)$ is the number of windows that occur exactly once, then:
\begin{equation}
R_k(t) \;=\; \frac{W_k(t)-S_k(t)}{W_k(t)}.
\end{equation}

To compare independent runs, we convert $R_k(t)$ in a binary criterion: 

\begin{equation}
E_k(t) \;=\; \mathbb{1}\!\left[R_k(t)\ge \tau\right].
\end{equation}

The plotted curve  (Fig. \ref{fig:Motifs}) is then:
\begin{equation}
P(k,t) \;=\; \frac{1}{N}\sum_{i=1}^{N} E^{(i)}_k(t),
\end{equation}
the fraction of simulation runs (out of $N$ seeds) that satisfy the repetition criterion at cycle $t$. We use $\tau=0.10$ and report 95\% Wilson confidence intervals for the estimated proportion.

\subsubsection{Complementary association increases the prevalence of repeated longer motifs}
\begin{figure}[H]
    \centering
    \includegraphics[width=1.05\textwidth]{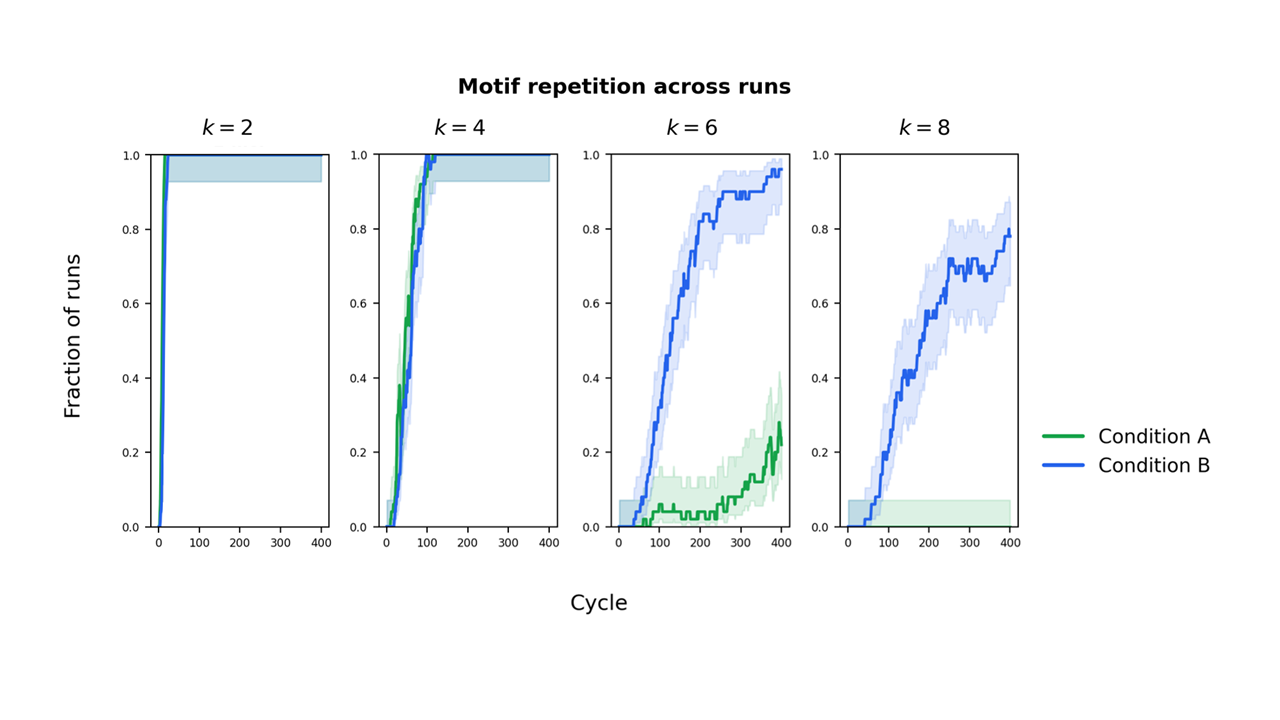}
    \caption{\textbf{Motif repetition across runs.}
    For each cycle $t$ and motif length $k$, we measure the fraction $R_k(t)$ of length-$k$ windows that correspond to motifs observed more than once in the population.
    A run is counted as positive if $R_k(t)\ge\tau$ (here $\tau=0.10$).
    Curves show $P(k,t)$, the fraction of $N=50$ independent runs meeting this criterion, comparing Condition A (elongation-only; green) and Condition B (full DNA norms; blue).
    Shaded regions indicate 95\% Wilson confidence intervals.}
    \label{fig:Motifs}
\end{figure}

For short motifs ($k=2$--$4$) (Fig. \ref{fig:Motifs}), the repetition criterion is met rapidly in both conditions: most runs quickly reach a state where at least 10\% of $k$-mer windows are repeats. For longer motifs (Fig. \ref{fig:Motifs}), the conditions diverge. Under Condition B (full DNA norms), many runs satisfy the same criterion for $k=6$--$8$ within the simulated time window, whereas under Condition A (elongation-only) runs rarely reach the threshold for these longer lengths.

A natural interpretation is that complementary association (including overlap-mediated merging) facilitates the formation of longer duplex-like regions, which can then split to yield two single-stranded offspring. This increases the reuse and duplication of longer contiguous sequence segments and therefore increases the fraction of $k$-mer windows occupied by repeated motifs at larger $k$.

\subsubsection{Spatial organization and local sequence heterogeneity}
\begin{figure}[H]
    \centering
    \includegraphics[width=1.05\textwidth]{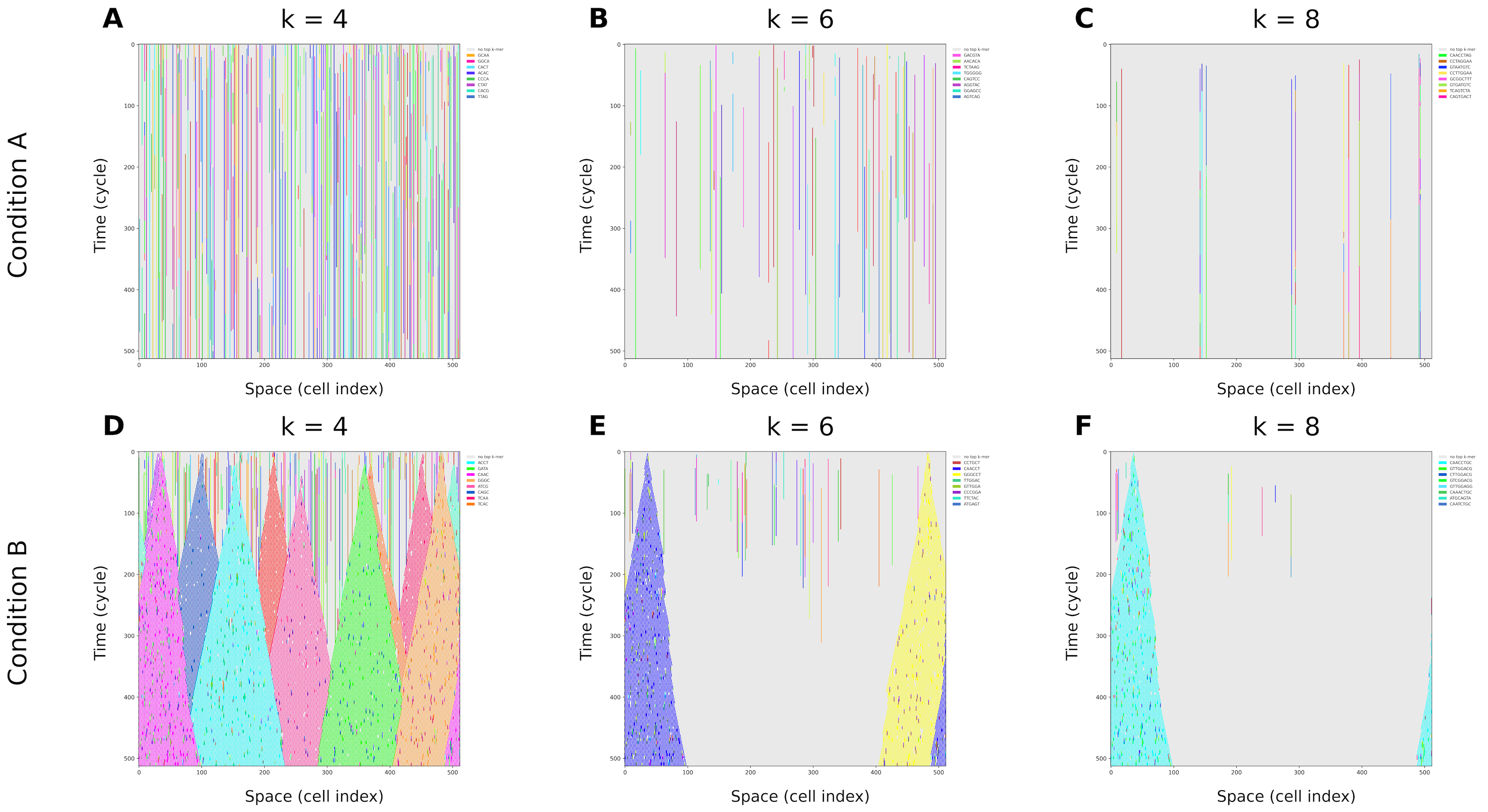}
    \caption{\textbf{Spatial signature of DNA-norms in a 1D cellular automaton.} Spacetime diagrams (space on $x$, cycle on $y$) show the dominant $k$-mer identity per cell at each cycle for elongation-only (Condition~A; panels A--C) and full DNA-norms with complementary association/overlap-mediated merging and splitting (Condition~B; panels D--F). Columns correspond to motif length $k\in\{4,6,8\}$. Colors indicate the most frequent $k$-mers (legend in each panel); gray denotes no assignment to the displayed top set. Under Condition~A, dominance appears primarily as sparse, short-lived streaks that rarely spread, becoming increasingly rare with larger $k$. Under Condition~B, dominant motifs form contiguous cone-shaped domains that expand from nucleation sites and persist over many cycles, remaining detectable even at $k=8$.
    }
    \label{fig:ca-spacetime}
\end{figure}

In Condition~A, the CA does not develop persistent replication or spatial structure (Fig.~\ref{fig:ca-spacetime}A--C). Dominant $k$-mers appear as thin, short-lived streaks. These motifs do not spread to neighboring sites, and the signal weakens as $k$ increases (Fig.~\ref{fig:ca-spacetime}B--C). In this regime, elongation can create sequence material, but without complementary association it does not reliably assemble longer strands that can reproduce.
By contrast, when complementary association (including overlap-mediated merging) and splitting are enabled (Condition~B), the CA exhibits clear spatial structure (Fig.~\ref{fig:ca-spacetime}D--F). Neighboring fragments can anneal through complementary overlaps, forming longer double strands; once sufficiently complete, these duplexes split and place offspring into adjacent cells. Repeating this local cycle produces laterally propagating offspring. Domains emerge for short motifs ($k=4$) and remain detectable for longer motifs ($k=6$ and even $k=8$), indicating that long contiguous sequence can be maintained and inherited under strictly local interactions.

A key qualitative outcome of Condition~B is the degree of heterogeneity. Even with only four nucleotide types and a minimal set of DNA-norm operators, multiple distinct motif identities coexist across the lattice (colors/legends in Fig.~\ref{fig:ca-spacetime}D--F). Different domains are dominated by different $k$-mers, and these domains nucleate, expand, and interact. Thus, the full DNA-norm generates heritable spatial structure, and sustains a diverse, spatially mixed population of sequence lineages from a minimal alphabet and minimal local rules.

\subsubsection{Future expansion}
Our minimal implementation isolates the contribution of the DNA-norms to motif reuse and spatial lineage formation. Several natural extensions would move it closer to Barricelli's broader program of open-ended symbiogenesis.
A first step is to introduce explicit sequence-dependent operational structures (e.g., binding sites, catalysts) that act back on the environment, for instance by modulating local resource availability. This would allow genomes to influence replication indirectly through functional components. A second extension would be to assign to the DNA system a genetic language so that sequences would encode rules for assembling multi-part structures, or for recruiting and arranging components into higher-order complexes, making symbiotic integration a mechanism for constructing composite organisms. Third, richer physical constraints could replace the current schematic implementation of stability and completion with explicit energy budgets and kinetic costs. Fourth, it would be interesting to include explicit recombination mechanisms such as strand breakage and ligation, multi-template complexes, or template switching during completion. Finally, it is natural to ask whether parts of the machinery implementing the DNA-norms could themselves be influenced by sequence context, raising the broader question of co-adaptation between genomes and their interpreter; related perspectives are discussed in Section~\ref{subsec:interpreter}. Moving from DNA-like pairing to an explicit RNA regime would likewise connect the present norms to the computational perspective in Section~\ref{subsec:rna}.

\section{Discussion and Implications}    
\subsection{Related Substrates}
Ideal substrates to test symbiogenesis are Neural Networks and Neural Cellular Automata (NCAs) \citep{Hartl2026, Mordvintsev2020}, but also other collective systems, and different languages as BFF \citep{alakuijala2024computational}.

\subsection{Implications for AI and Alife}
\begin{itemize}
    \item \textbf{Limitations of SOS search algorithms.}
    Many existing “symbiotic” optimization approaches (e.g., Symbiotic Organism Search) treat symbiosis as a heuristic for cooperative optimization rather than as an emergent mechanism that reorganizes the reproductive unit of evolving systems.
    \item \textbf{Symbiotic evolutionary algorithms.}
    Further research is needed on evolutionary algorithms where symbiotic associations between genomes or programs can emerge dynamically and form new composite individuals.
    \item \textbf{Evolvability and collective intelligence.}
    Symbiogenesis may increase evolvability by enabling systems to combine independently evolved functional components into cooperative collectives, thereby driving evolutionary innovation~\citep{Wagner2011evoinnovation, Kauffman1993originsoforder}. This process is notably akin to collective intelligences.
    \item \textbf{Open endedness.}
    Symbiogenesis may promote evolvability by enabling systematic abstraction of complex systems across scales, into ever higher-order units of organization.
    \item \textbf{Connections to computation.}
    The true power of symbiogenetic might originate from compositional principles in theoretical computer science, such as combinatory logic and lambda calculus~\citep{akhlaghpour2022rna}.
\end{itemize}

\subsection{Co-Evolving Code and Interpreter: A Missing Link for Environmental Feedback?}
\label{subsec:interpreter}
The standard view of genetics holds that genes encode phenotypical traits of organisms, and that selection acts on those traits. Reproduction through a genetic bottleneck introduces variation via crossover and mutation, allowing beneficial genes to persist while unfavourable ones perish, thereby driving adaptation and evolution through random mutation and selection. A more accurate picture, however, is that genes primarily encode protein sequences (among other functional elements). Once transcribed, proteins become nano-machines that operate in the context of their cellular environment and continuously reconfigure their host system~\cite{Levin2023DAM}.

Biology is working with messy hardware and software. To achieve stable and canalizing outcomes during organism development and maintenance across scales, the puzzle pieces of biological organization--from cells, tissue, organoids, towards organisms, and possible beyond--must actively navigate their own problem spaces \citep{Fields2020ScaleFree,Fields2022Competency}. This implies the presence of an intermediate computational layer between the genetic encoding of an organism and its anatomical form and function, as discussed for instance in the framework of multiscsale competency architectures \citep{Levin2023DAM}. Evolution operates on an agential material: the components of living systems can deal with novel situations on the fly (as illustrated by galls, Xenobots, or regenerative experiments), which can in turn accelerate evolvability and facilitate transfer learning across evolutionary timescales \citep{Hartl2024MCA}. In this perspective, evolution does not merely search for the most fitted organisms, but brings forth problem-solving entities capable of integrating into collectives of collectives across scales of biological organization. If we adopt the view of biology as a multiscale competency architecture--or, in modern terminology, a form of collective intelligence--then the genomic bottleneck, together with gene regulatory and developmental mechanisms, can be understood as a functional problem-solving unit, describable as a generative model operating within its multicellular environment \citep{mitchell2024genomic,hartl2025generativegenome}.

\textbf{Why is this relevant in the context of symbiogenetic (computational) systems?}

An evolving genetic code alone--even a rule-set for replication is present--is only half of the story. Genetic sequences are interpreted in the context of the cellular host: coding blocks are transcribed into functional units, such as proteins, that interact with their biochemical environment. The ribosome acts as a molecular decoder, yet the broader cellular context--including cytoplasmic conditions, regulatory networks, interactions with other cells, and environmental signals and developmental stages--effectively determines how the genetic code is interpreted at any given moment. This interpretation layer displays active problem-solving properties. For example, regenerative experiments with planarian flatworms placed in Barium solution show that cells can identify and express the correct subset of genes--among tens of thousands--to rebuild a head that remains stable under these novel conditions~\citep{durant2017long}. This is remarkable since Barium has with large certainty never been a factor in their evolutionary history: this adaptation to barium resistance happens within the same specimen. In this sense, gene regulation, readout, and replication are not passive processes but active, context-dependent computations.

Recent work in developmental biology, artificial life, and machine learning increasingly points toward similar architectural principles. In biological systems, the genome does not function as a static blueprint but rather as a compact encoding that unfolds through distributed regulatory dynamics across cells and tissues~\citep{hartl2025generativegenome,mitchell2024genomic}: Cellular collectives interpret and reconfigure gene expression patterns through local signaling, feedback, and environmental interactions, effectively implementing a distributed developmental computation. Related ideas have also emerged in artificial systems, where adaptive models construct internal representations or embedding spaces that are dynamically remapped as new information becomes available~\citep{Hartl2026ReNAS}. 
In both cases, interpretation is not merely the decoding of instructions but an active process that maps encoded information into functional behavior under changing environmental conditions and co-evolving internal dynamics.

From this perspective, not only the genetic code evolves, but also the engine that interprets it--potentially across multiple spatial- and timescales spanning evolution, development, and learning. Changes in this interpreter modify not only the content written on the tape, but also the language by which that tape is read. In most artificial life and computational evolution systems, however, the interpreter is fixed. For instance, the BFF replicator experiments in computational life \citep{alakuijala2024computational} assume a predefined programming language and compiler, while Barricelli’s experiments rely on fixed interaction norms. Many physical substrates--such as chemical reaction networks--naturally provide rule sets that can implement computational processes. Yet the emergence of life-like organization appears to depend less on the mere existence of such rules and more on the particular combinations that allow them to interact productively. In other words, the right combination makes the soup. This suggests that the interpreter itself may need to co-evolve with the evolving system: life achieves robust functionality using unreliable hardware and software precisely because the genomic code and its interpreter jointly evolve toward mutually beneficial robustness and adaptability in the face of novelty.

\subsection{From environmental constraints to cooperation through combinatorial functional niching}
Our biosphere exists in a resource-constrained environment, which naturally leads to competition between organisms. At the same time, organisms must persist within the conditions imposed by their environment. One promising formalization of this requirement is given by the free energy principle and active inference frameworks, in which persistence corresponds to minimizing surprise or prediction error with respect to environmental states \citep{Friston2010, Friston2017}. In practice, this can be achieved in two complementary ways: organisms can act to modify their environment so that it becomes more predictable, or they can update internal belief states about the environment through learning~\cite{Pezzulo2016}. Survival therefore requires not only learning how to act, but also learning the relevant modalities of the surrounding environment.

Importantly, organisms do not merely react to their environment; they can also utilize and reshape parts of it for their own benefit. Their behavior is therefore conditional on environmental structure and constraints \citep{Fields2022Competency, Fields2020ScaleFree, Hartl2026ReNAS}. Different environmental constraints may reward different sets of capabilities, leading to diverse evolutionary strategies and skill sets. A natural question then arises: what happens when organisms with different capabilities coexist or compete within the same environment? Game-theoretic results suggest that, over long time horizons, collaborative strategies can outperform purely adversarial ones, as illustrated by experiments on the iterative prisoner's dilemma~\citep{Axelrod1981}. In this context, symbiotic interactions can become an advantageous strategy, allowing collective agents to distribute labor and specialize in complementary functions. Computational studies have suggested that such symbiotic composition can increase evolvability by enabling systems to combine independently evolved functional modules into mutually beneficial higher-order structures~\citep{watson2001symbiotic, watson2003computational}.

Individuality is dynamic: symbiotic associations can ultimately lead to evolutionary transitions in individuality, where previously independent entities merge into novel individuals that can no longer exist independently. Such higher-order organizational units may capitalize on the functional resources of their constituent parts~\citep{Fields2022Competency,Fields2020ScaleFree,Hartl2026ReNAS} and utilize them ``on demand'' in a combinatorial manner. We refer to this principle as ``\textit{combinatorial functional niching}'': the ability of a collective system to combine and integrate orthogonally acquired functionalities through specialization and cooperation to achieve new functional capabilities \citep{watson2001symbiotic,watson2003computational}. In principle, this mechanism allows collaborative hierarchical collectives--or multiscale competency architectures--to access a combinatorially expanding space of capabilities without requiring every component to independently evolve all necessary functions.

This perspective is particularly intriguing at the genetic and regulatory level. Gene regulatory networks (GRNs) can address genes flexibly and context-dependently, enabling combinations of gene products that support different physiological or developmental outcomes. Following symbiogenetic events, such regulatory mechanisms may allow the integrated system to reason over newly combined genetic repertoires and deploy them in novel contexts. In this sense, symbiogenesis may not only increase genetic diversity but also expand the functional space that a system can explore, potentially providing a probabilistic advantage for persistence within a complex and highly entropic environment.

Combinatorial functional niching might be tested in environments that require multiple difficult-to-learn functions. Different organisms could specialize in different tasks and cooperate as a collective through symbiotic interactions. Symbiogenetic events may then allow these capabilities to be combined, enabling higher-level control structuress to coordinate integrated functionality. In such a setting, an interpreter that can access and orchestrate the functional repertoire of multiple symbionts could merge their capabilities to gain a combinatorial advantage.

Several related questions arise from this perspective:
\begin{itemize}
    \item We know that niche construction exists $\rightarrow$ Is symbiogenesis a promoter of niching?
    \item Is symbiogenesis an ingredient for kickstarting life, vs. promoting open-endedness?
    \item Are symbiosis and symbiogenesis the same thing, but represent different stages of a collaborative integration process?
    \item Considering higher levels of organization beyond genes, combinatorial functional niching may provide an organizing principle for evolutionary transitions and shifts in the reproductive unit. It raises the possibility that reproductive units can actively reorganize to explore niches more efficiently.
    \item Considering cultural evolution, ``from genes to memes'': How do minds, companies, or societies "reproduce"? Is learning/acquiring new memories a symbiogenetic event for our minds~\citep{hartl2025generativegenome, levin2024selfimprovisingmemory}? Do societies age~\citep{PioLopez2025}?
\end{itemize}

This suggests that symbiogenesis may not only generate new organisms, but may also expand the functional niche space accessible to evolving systems across scales more generally.
We close this section with a final thought: Does the interpreter exist independently of the genome, or is it itself another symbiogenetic system that keeps this dual engine of life running (c.f. the relationship between genomes, and proteins, cytoplasmic machinery)? 

\subsection{RNA: A Physical Turing-Complete Language}
\label{subsec:rna}
Recent work has suggested that RNA can act as a computational substrate capable of implementing a language with the same expressive power as a universal Turing machine~\citep{akhlaghpour2022rna}. In this work, the authors relate and define combinatory logic operators (such as the classical B, C, K, and W combinators) with biochemical mechanisms including base pairing, strand displacement, and splicing-like transformations of RNA strands. This framework positions RNA molecules and their interactions as a physical system capable of executing combinatory programs. Structurally, such computations are realized through the formation of loops, junctions, and matching addresses along RNA strands, effectively creating a dynamically reconfigurable molecular program tape; see also Sections~\ref{sub:DNA:norm} and \ref{sub:DNA:results}.

This observation raises interesting questions for artificial life systems inspired by early computational evolution experiments. For example, Barricelli’s automata operate through collision-based interaction norms on a one-dimensional lattice. These norms determine how numerical sequences replicate and interact when they encounter one another. While these rules are capable of generating self-reproducing structures and symbiogenetic events, it remains unclear--in contrast to Turing-complete languages such as BFF~\citep{alakuijala2024computational}--whether the underlying rule set is computationally universal. In particular, Barricelli’s system does not appear to naturally support the formation of persistent addressing mechanisms or recursive control structures analogous to those found in universal computational systems.

This leads to a natural question: what minimal rules or interaction norms would be required for a one-dimensional automaton to become computationally universal in an RNA-like sense? In particular, would the emergence of universal computation require the spontaneous formation of addressing mechanisms, loops, or other compositional structures that resemble symbiotic integration between interacting programs?

From a biological perspective, this possibility is intriguing. RNA and DNA exist within a thermodynamic environment composed of large combinatorial spaces of interacting molecules. Surviving in such an environment may require systems to reason over combinations of molecular interactions and gene expression patterns. In this sense, the regulatory and interpretive machinery of living systems may implicitly implement computations over combinatorial spaces of molecular states. If so, the molecular languages of life may naturally approximate computational systems capable of universal computation, implemented directly within biochemical substrates.

Understanding which minimal interaction rules allow such computational capacities to emerge could therefore help bridge the gap between early artificial life models, such as Barricelli’s symbioorganisms, and modern views of biological computation.

\subsection{Ideas for future work}
\begin{itemize}
    \item \textbf{Task environments for symbioorganisms.}
    Barricelli originally introduced structured environments by allowing symbioorganisms to play the game Tac Tix. Extending this idea, one could design environments containing multiple tasks or ``games''. Different organisms could specialize in different tasks, while symbiotic associations allow cooperative strategies to emerge. Symbiogenetic events could then integrate these specialized capabilities into higher-level organisms that can exploit multiple strategies simultaneously.
    \item \textbf{Evolving interpreters.}
    A central conceptual question concerns the nature of the interpreter. In most artificial life systems, the rules governing interactions are fixed and effectively act as the ``laws of the universe''. Biological systems, however, rely on complex interpretive machinery that translates genomic information into functional behavior. In this context, even the existence of a stable rule-set is a non-trivial presumption: both the code and its interpreter must ultimately be implemented by physical processes. The laws of physics do not inherently perform computation, yet living systems have evolved ways to utilize physical states/substrates to encode and process information \citep{Wolpert2026}. This raises intriguing questions: do interpreters themselves need to evolve? Could symbioorganisms influence the mechanisms that interpret their genomes, thereby increasing their own viability or evolvability?
    \item \textbf{computational testbed} One possible computational testbed for these ideas could involve evolutionary experiments that explicitly decouple genomic code from its interpreter. For instance, the genome could be represented as a sequence of symbols, while a learned attention mechanism (e.g., a QKV-attention head) acts as the interpreter that reads and transforms this sequence. In such a setting, genes or blocks of genes could attend to arbitrary positions in the sequence, enabling flexible recombination and functional composition, analogous to the combinatorial functional niching discussed above. An open question is how the parameters of the interpreter (e.g., the attention mechanism) could co-evolve with the genome itself, possibly via indirect or developmental encodings.
    \item \textbf{Emergent biological functionality.}
    Can sensing mechanisms emerge in such systems? Can membranes or compartmentalization arise as a consequence of evolutionary pressures? More generally, can interpreters themselves emerge, potentially giving rise to self-coding or self-modifying languages?
    \item \textbf{Computational universality.}
    Are any of Barricelli's interaction norms computationally universal? More generally, what minimal rule sets or interaction norms are sufficient to support universal computation in evolutionary automata? Investigating such questions could clarify the relationship between replication dynamics, symbiogenesis, and the emergence of computationally expressive systems.
    \item \textbf{Connections to mathematics and theoretical computer science.}
    Are there analogues of replicators, loops, or symbiogenetic integration in formal systems such as lambda calculus, combinatory logic, or proof systems? Exploring such connections may help reveal deeper structural principles underlying replication and composition in computational systems.
\end{itemize}

\section{Conclusion}
\textbf{It's complicated!! But fun :)}

This report explored symbiogenesis from three complementary perspectives. 
First, we revisited Barricelli’s numerical symbioorganisms and implemented several computational substrates inspired by his framework, including 1D and 2D cellular automata as well as a minimal Boolean CA variant to explore the simplest settings in which symbiogenetic dynamics may arise. 
Second, we introduced minimal DNA-inspired interaction norms to investigate how molecular-like local rules can give rise to replication, sequence reuse, and spatial lineage formation. 
Third, we discussed broader implications of symbiogenesis for artificial life and artificial intelligence, including collective intelligence, evolving interpreters, and combinatorial functional organization in computational systems.

Note: This \textit{report} is the result of a 1-week group work at the ALICE 2026 workshop. It is not meant to be a properly written scientific paper, rather a collection of ideas, thoughts, preliminary results, and considerations that emerged during brainstorming and group discussions. Revisiting Barricelli's work has provided many open directions for future work, many of which are discussed (altough often in bullet-point lists) in this document. We call the ALife community to revise and expand of Barricelli's pioneering ideas and we hope this report motivates other colleagues to investigate further. If ideas in this report are pursued further or expanded, we kindly ask to cite this report.

\section*{Data and code policy}
All code for the 1D/2D Barricelli systems is available \href{https://www.github.com/JELAshford/symba-alice-2026}{here}

\section*{Acknowledgments}

We would like to thank the ALICE 2026 organizers. Additionally, special thanks to Tim Taylor for copies of the original manuscripts from Barricelli, Luca Manzoni for useful material from a keynote talk at SSCI, and Kai Olav Ellefsen for scanned copies of manuscripts from the Oslo University library.

\printbibliography
\end{document}